\pdfoutput=1

\documentclass[11pt]{article}

\usepackage[final]{acl}

\usepackage{tcolorbox}
\tcbuselibrary{listings}
\tcbuselibrary{breakable}
\usepackage{times}
\usepackage{latexsym}
\usepackage{booktabs,multirow}
\usepackage{siunitx}
\usepackage{geometry}
\usepackage{amssymb}
\usepackage{xcolor}
\usepackage{listings}
\usepackage{tablefootnote}

\usepackage{enumitem}
\usepackage[T1]{fontenc}

\usepackage[utf8]{inputenc}

\usepackage{microtype}

\usepackage{inconsolata}

\usepackage{graphicx}

\title{From Scores to Steps: Diagnosing and Improving LLM Performance in Evidence-Based Medical Calculations}

\author{
Benlu Wang\thanks{Equal contribution, alphabetical order}~$^{1}$, 
Iris Xia\footnotemark[1]~$^{1}$, 
Yifan Zhang~$^{2,3}$,
Junda Wang~$^{2,4}$, \\
\textbf{Feiyun Ouyang}~$^{2,3}$,
\textbf{Shuo Han}~$^{3}$,
\textbf{Arman Cohan}~$^{1}$,
\textbf{Hong Yu}\thanks{Co-corresponding authors}~$^{2,3,4}$,
\textbf{Zonghai Yao}\footnotemark[2]~$^{2,4}$ \\
$^{1}$Department of Computer Science, Yale University, CT, USA \\
$^{2}$Center for Healthcare Organization and Implementation Research, VA Bedford Health Care  \\
$^{3}$Miner School of Computer and Information Sciences, UMass Lowell, MA, USA \\
$^{4}$Manning College of Information and Computer Sciences, UMass Amherst, MA, USA \\
\texttt{\href{mailto:benlu.wang@yale.edu}{benlu.wang@yale.edu}, \href{mailto:iris.xia@yale.edu}{iris.xia@yale.edu}, \href{mailto:zonghaiyao@umass.edu}{zonghaiyao@umass.edu}}
}

\begin{document}
\maketitle
\begin{abstract}

Large language models (LLMs) have demonstrated promising performance on medical benchmarks; however, their ability to perform medical calculations, a crucial aspect of clinical decision-making, remains underexplored and poorly evaluated. 
Existing benchmarks often assess only the final answer with a wide numerical tolerance, overlooking systematic reasoning failures and potentially causing serious clinical misjudgments.
In this work, we revisit medical calculation evaluation with a stronger focus on clinical trustworthiness. First, we clean and restructure the MedCalc-Bench dataset and propose a new step-by-step evaluation pipeline that independently assesses formula selection, entity extraction, and arithmetic computation. Under this granular framework, the accuracy of GPT-4o drops from 62.7\% to 43.6\%, revealing errors masked by prior evaluations.
Second, we introduce an automatic error analysis framework that generates structured attribution for each failure mode. Human evaluation confirms its alignment with expert judgment, enabling scalable and explainable diagnostics.
Finally, we propose a modular agentic pipeline, MedRaC, that combines retrieval-augmented generation and Python-based code execution. 
Without any fine-tuning, MedRaC improves the accuracy of different LLMs from 16.35\% up to 53.19\%. 
Our work highlights the limitations of current benchmark practices and proposes a more clinically faithful methodology. 
By enabling transparent and transferable reasoning evaluation, we move closer to making LLM-based systems trustworthy for real-world medical applications.

\end{abstract}

\section{Introduction}

Clinical calculation matters, but benchmarks miss the point.
While large language models (LLMs) are increasingly used in clinical settings~\cite{achiam2023gpt,goodman2023accuracy,decker2023large,ayers2023comparing,thirunavukarasu2023trialling,singhal2023large,sun2024effectiveness,yao2025survey} for question answering~\cite{app11146421}, medical documentation summarization~\cite{shaib2023summarizing}, and even decision support~\cite{Thirunavukarasu2023,tu2025towards}, many of these applications hinge on the model’s ability to perform reliable medical calculations~\cite{goodell2025large}. 
Such tasks, like computing glomerular filtration rate or cardiovascular risk, require high numerical accuracy, correct formula use, and context-aware data extraction~\cite{cockcroft1976prediction,initiative20102010,gage2001validation}. 
Yet existing benchmarks for evaluating LLMs in this domain fall short of these requirements.


MedCalc-Bench~\cite{NEURIPS2024_99e81750} recently introduced a collection of real-world medical calculation tasks, drawn from widely used calculators that surveys show were regularly used by over $80\%$ of healthcare professionals today~\cite{MDCalcAboutUs}.
However, its current evaluation protocol only checks whether the final answer falls within a ±5\% tolerance.
This overlooks critical failures in intermediate steps, such as selecting the wrong formula, misreading patient attributes, or miscalculating values, creating an illusion of high performance while masking real risks.
Additionally, we observed corrupted data points that hindered the analysis of model performance.

We address this problem by first cleaning errors in the original benchmark and proposing a three-part framework for more faithful evaluation and performance enhancement:

\begin{itemize}[leftmargin=.1in,topsep=0.3pt]
\setlength\itemsep{0em}
    \item A step-by-step evaluation pipeline that assesses each reasoning stage: formula selection, entity extraction, and numerical computation.~\footnote{Our code and data are released here: \url{https://github.com/Super-Billy/EMNLP-2025-MedRaC} with Apache-2.0 license.}
    \item An LLM-based automatic error analysis framework that attributes mistakes to specific steps and generates structured explanations, validated against human experts.
    \item A training-free, agentic enhancement method that decomposes medical calculation into distinct stages and leverages retrieval-augmented grounding with executable code generation to reduce hallucinations and boost accuracy.
\end{itemize}

While many recent benchmarks report steady gains in accuracy, it remains unclear how much these improvements translate into safer or more deployable systems in high-stakes domains~\cite{kung2023performance,jin2024hidden,yang2025unveiling}. 
By rethinking how we evaluate and support LLMs in medical calculation, an essential, tool-heavy aspect of real clinical practice, we offer a more transferable and trustworthy pathway from NLP progress to clinical impact.

\section{Background and Related Work}

\paragraph{Limitations of Final-Answer Medical Benchmarks}
Early benchmarks, such as MedQA~\cite{app11146421}, PubMedQA~\cite{jin-etal-2019-pubmedqa}, and MedMCQA~\cite{pmlr-v174-pal22a}, primarily focus on factual recall and multiple-choice question answering. However, these benchmarks do not test models' ability to perform quantitative or step-by-step reasoning. 
MedIQ introduces a question-asking dataset that encourages models to seek missing information, though this often degrades performance~\cite{li2024mediq}.
MedCalc-Bench~\cite{NEURIPS2024_99e81750} improves upon this by introducing real-world medical calculation tasks. It draws from 55 widely used MDCalc calculators and includes 1047 patient vignettes covering scenarios such as glomerular filtration rate (GFR) estimation and body mass index (BMI) calculation. These tasks require selecting the correct formula, extracting clinical variables, and performing numerical computations. 
Despite its innovation, MedCalc-Bench only evaluates the final numeric answer, allowing a ±5\% margin of error. This can obscure errors such as formula misapplication, omission of key patient factors, or hallucinated arithmetic. As our reanalysis reveals, many answers marked as “correct” under the original metric contain faulty reasoning chains, thereby limiting their clinical reliability.
We extend MedCalc-Bench by introducing a step-wise evaluation pipeline that inspects each reasoning component—formula, extraction, computation, and answer formatting—independently, revealing deeper reasoning failures that would otherwise go undetected.

\paragraph{Evaluating Intermediate Reasoning with LLM-as-Judge}
Step-wise evaluation has gained traction in general NLP tasks~\cite{lightman2023let, chen2022pot, lee2025evaluating,shen2025let}, with LLMs increasingly used as automated judges~\cite{li2024generation,gu2024survey}. 
Studies show that models like GPT-4~\cite{achiam2023gpt, liu2023gpteval, fu2023gptscore} and critique-tuned variants~\cite{ke2023critiquellm} can approximate human judgment in summarization~\cite{chen2023storyer}, dialogue~\cite{zheng2023judging, zhang2024comprehensive}, and translation~\cite{kocmi2023large}.
In the medical domain, LLM-as-judge has been applied to clinical conversations~\cite{tu2025towards,arora2025healthbench,wang2023notechat}, medical documentation~\cite{croxford2025automating,chung2025verifact,brake2024comparing}, exam question answering \& generation~\cite{yao2024mcqg,yao2024medqa}, and medical reasoning~\cite{jeong2024improving,tran2024rare}.
Inspired by these works, we introduce the first step-wise LLM-as-Judge framework for clinical calculation tasks. 

\paragraph{Retrieval-Augmented and Execution-Based Enhancements}
In clinical NLP, hallucinations are a key concern, particularly in high-stakes applications. Retrieval-augmented generation methods~\cite{nori2023can,xiong2024benchmarking,xiong2024improving,wang2024jmlr} address this by grounding generation in trusted sources. 
Visual RAG approaches further improve reliability in imaging tasks~\cite{chu2025vrag}. Surveys confirm that RAG systematically reduces fabrication~\cite{zhang2025hallucination, amugongo2025retrieval}.
Program-aided reasoning and broader tool-use approaches have been extensively studied in recent work, highlighting the value of integrating external tools into LLM workflows~\cite{mialon2023augmented, pmlr-v202-gao23f}.
Parallel to retrieval, execution-based techniques like Self-Consistency~\cite{wang2022selfconsistency} and Self-Refine~\cite{madaan2023selfrefine} offer tools for reducing arithmetic and logical errors. These methods are often applied in math and symbolic reasoning, but have not been widely tested in clinical computations.
We unify both strategies into a plug-and-play agentic pipeline tailored to medical calculations. By combining formula retrieval with Python code execution, our method corrects both hallucination-driven and computation-driven errors—without requiring any model fine-tuning.

\begin{figure*}[ht]
  \centering
  \includegraphics[width=\textwidth]{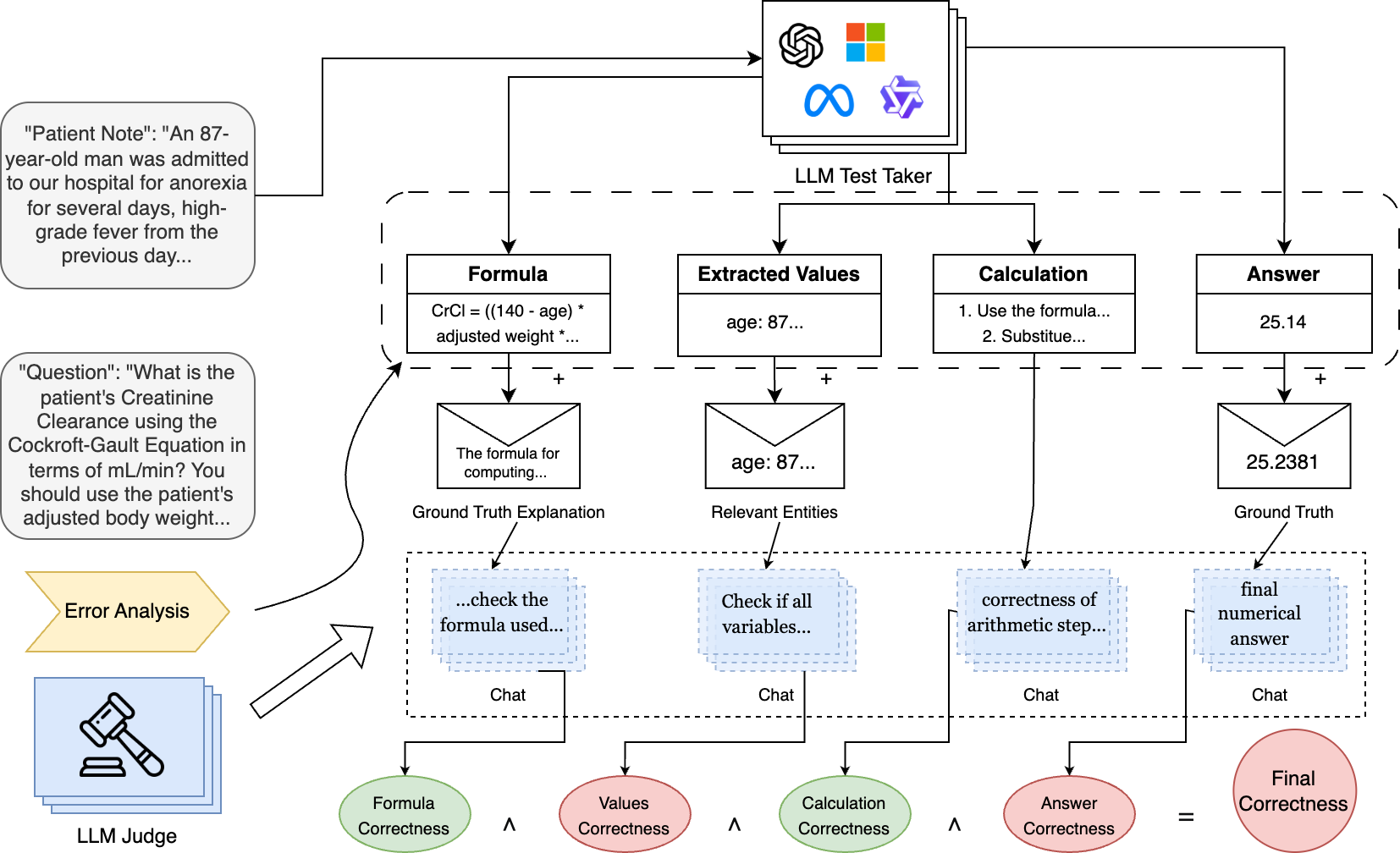}
  \caption{Step-wise LLM-aided Evaluation Pipeline. Each reasoning stage is checked by an LLM-Judge against a reference explanation to determine its correctness.}
  \label{fig:llm_evaluation}
\end{figure*}

\section{Methods}

\subsection{Step-wise Evaluation}
\label{Step-wise Evaluation}

As shown in Figure~\ref{fig:llm_evaluation}, medical calculations typically involve multiple sequential steps, such as retrieving relevant medical knowledge and identifying the appropriate formula.
We propose a structured evaluation pipeline for medical calculation tasks that decomposes the reasoning process into four sequential, individually validated steps:

\paragraph{Formula selection.}  
    The candidate response must employ the \emph{correct} medical calculation formula, as defined among the 55 calculators in MedCalc-Bench, and specify it fully, including appropriate units, boundary conditions, and any relevant constraints. We constructed a reference formula library corresponding to these 55 calculators, against which each model-proposed formula is evaluated. We use an evaluator to compare the predicted formula to its canonical counterpart in this library and assign a binary correctness score.
    
\paragraph{Value extraction.}  
    We ask the evaluator to extract every numerical and categorical variable from both the clinical vignette and the model’s response. These extracted variables are then compared against the gold-standard answers provided in the dataset's JSON annotations. Using a closed-book LLM evaluator, we compute the alignment and assign a binary correctness score: full agreement is required to pass, while any mismatch, such as a missing, hallucinated, or incorrectly labeled variable, results in failure.
    
\paragraph{Mathematical calculation.}  
    The evaluator verifies whether each arithmetic step is valid, based on the extracted formula and values. UUnlike MedCalc-Bench, which allows a 5\% margin of error, we adopt a stricter criterion, following the tolerance defined on the original calculators’ website, MDCalc. Specifically, the allowed numerical tolerance depends on the number of decimal places in the LLM’s answer, capped at two decimal places. For example, an answer of 10.65 is evaluated with a $\pm 0.005$ tolerance, while answers with more than two decimals (e.g., 10.6512) are rounded and assessed with the same $\pm 0.005$ threshold. A binary correctness score is then assigned.

\paragraph{Final Answer.} 
    We evaluate whether the model’s final prediction is equivalent to the ground-truth answer in the dataset, allowing for valid unit conversions.

To ensure that the model focuses solely on evaluating the correctness of the mathematical computation, we provide only the LLM-generated answer as input, excluding the ground-truth answer or any reference formulas, to avoid potential bias or leakage that could influence judgment.

Let $\mathcal{S}_i$ denote the result of the $i$th step in the calculation process, each step is dependent on the previous steps: $\mathcal{S}_i = f(\mathcal{S}_{i-1}, \ldots, \mathcal{S}_1)$. Define a validation function $
\mathcal{V}(\cdot) \in \{\texttt{True}, \texttt{False}\}$. We propose that a step $\mathcal{S}_i$ can only possibly be correct if and only if the immediately preceding step is correct, that is,
\[
\Diamond \mathcal{V}(\mathcal{S}_i) \iff \mathcal{V}(\mathcal{S}_{i-1}).
\]
Our evaluation metric ensures correctness by verifying the validity of each step in the MedCalc Bench dataset. Specifically, we evaluate the following steps sequentially: formula correctness $\mathcal{V}(\text{F})$, extraction correctness $\mathcal{V}(\text{E})$, calculation correctness $\mathcal{V}(\text{C})$, and final answer correctness $\mathcal{V}(\text{A})$. We define the correctness of the calculation task for one case $\kappa \in \{\texttt{True}, \texttt{False}\}, $ as the conjunction of validity across all individual steps:

\[
\kappa = \mathcal{V}(\text{F}) \land \mathcal{V}(\text{E}) \land \mathcal{V}(\text{C}) \land \mathcal{V}(\text{A})
\]

Further, we define the \emph{Conditional Correctness} of each step $\mathcal{S}_i$ as the probability that the step is correct given that all preceding steps are correct:
\[
\text{CC}_i = \mathbb{P}\left(\mathcal{V}(\mathcal{S}_i) \mid \mathcal{V}(\mathcal{S}_1) \land \ldots \land \mathcal{V}(\mathcal{S}_{i-1}) \right).
\]
We also define the \emph{First Error Attribution Rate} of step $\mathcal{S}_i$ as the proportion of examples in which $\mathcal{S}_i$ is the first step to fail, i.e., all previous steps are correct but $\mathcal{S}_i$ is incorrect:
\[
\text{FE}_i = \mathbb{P}(\mathcal{V}(\mathcal{S}_1) \land \ldots \land \mathcal{V}(\mathcal{S}_{i-1}) \land \lnot \mathcal{V}(\mathcal{S}_{i}) \mid \lnot \kappa).
\]

This decomposition enables fine-grained error diagnosis and quantitative comparison across models and methods. 
Figure~\ref{fig:llm_evaluation} illustrates the whole pipeline.

\begin{figure}[!ht]
  \includegraphics[width=\columnwidth]{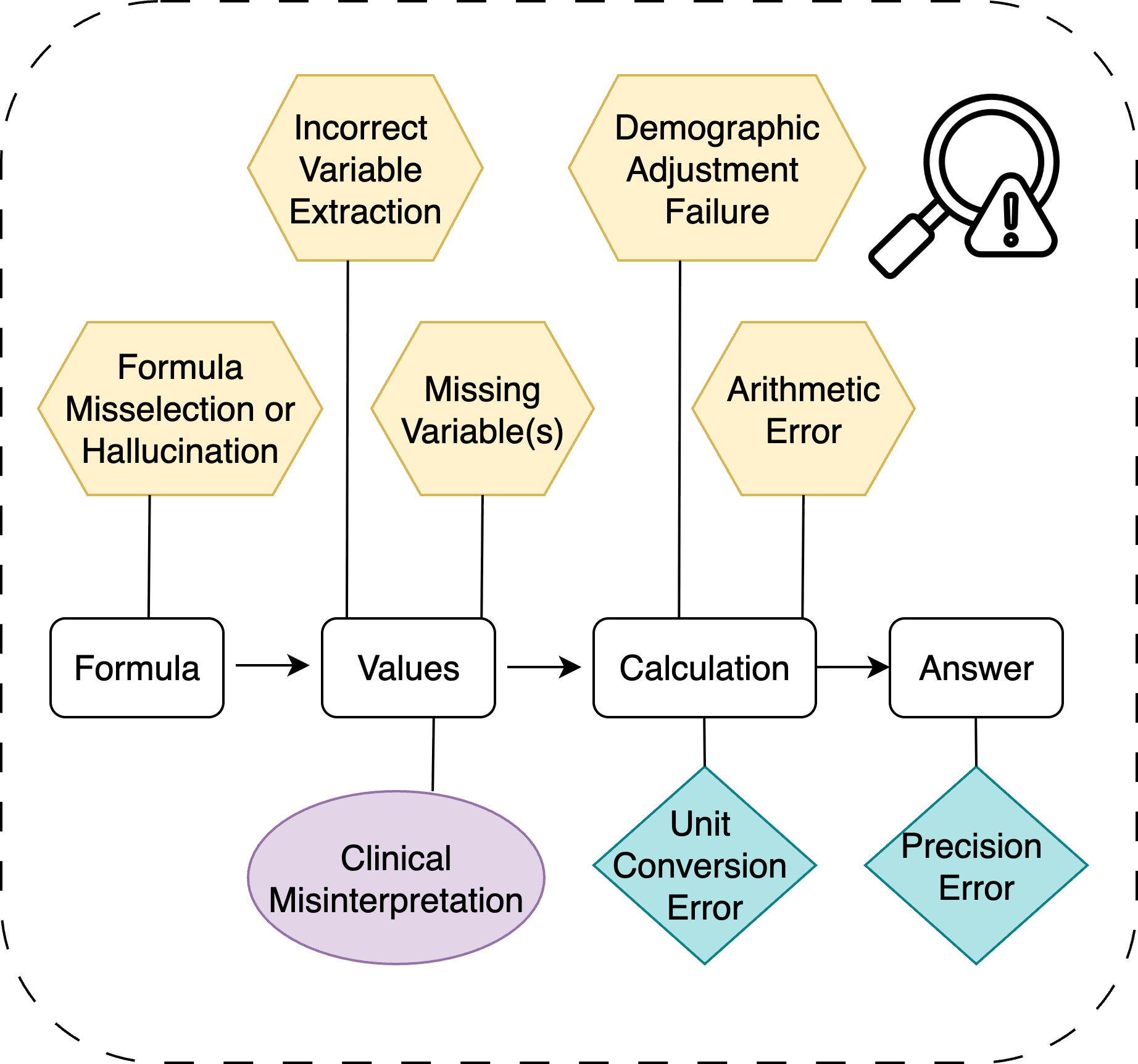}
  \caption{Categorization of Reasoning Errors in Clinical Calculation Tasks. Incorrect outputs can stem from diverse sources of failure across reasoning stages.
  }
  \label{fig:error_types}
\end{figure}

\subsection{LLM-aided Evaluation and Structured Error Attribution}

Building upon the step-wise evaluation pipeline, we design an LLM-aided judge to assess correctness at each stage. Given an input–output pair from the LLM Test Taker and a ground-truth reference (e.g., extracted variable, formula used, computed value), we prompt a high-performance LLM Judge to determine semantic alignment and provide binary correctness feedback.

To further analyze the failure patterns behind incorrect answers, we define a taxonomy of common medical calculation errors, visualized in Figure~\ref{fig:error_types}. Each failure is assigned to one or more of the following categories:

\textit{Formula Misselection or Hallucination:} The answer chooses a formula that does not fit the clinical scenario \emph{or} distorts the correct formula by inventing, omitting, or misplacing terms, coefficients, or operators (e.g.\ using Cockcroft–Gault instead of CKD-EPI for an AKI patient).

\textit{Incorrect Variable Extraction:} A wrong value, unit, or time-point is pulled from the note (e.g.\ yesterday’s creatinine, or treating \si{\micro\mole\per\liter} as \si{\milli\gram\per\deci\liter}).

\textit{Clinical Misinterpretation (Rule-based):} Numbers are captured correctly, but their clinical meaning is misjudged—wrong severity, threshold, or presence/absence decision (e.g.\ calling “trace ascites” “no ascites”).

\textit{Missing Variable(s):} One or more required inputs (weight, race, age group, etc.) are absent, yet the calculation proceeds, rendering the result unreliable.

\textit{Demographic Adjustment Failure:} A mandatory sex, race, BSA, pregnancy, or age multiplier is skipped or applied to the wrong group (e.g.\ omitting the 0.85 female factor).

\textit{Unit Conversion Error (Equation-based):} A value is used without the necessary unit change, or with an incorrect factor/direction, before substitution into the formula (e.g.\ using 134 \si{\micro\mole\per\liter} as 134 \si{\milli\gram\per\deci\liter}).

\textit{Arithmetic Error:} Pure math is wrong despite correct formula and inputs, basic addition, order of operations, exponentiation, or duplication/omission of terms.

\textit{Rounding / Precision Error (Equation-based):} The final number is outside the allowed tolerance solely because of over- or under-rounding (rule of 1–2 d.p.: ±0.05 for one decimal place, ±0.005 for two).

These error types enable structured analysis of model behavior and inform targeted interventions in later modules.

\subsection{MedRaC: Multi-Agent Enhancement with Formula-RAG and Code}

\begin{figure}[ht]
  \includegraphics[width=\linewidth]{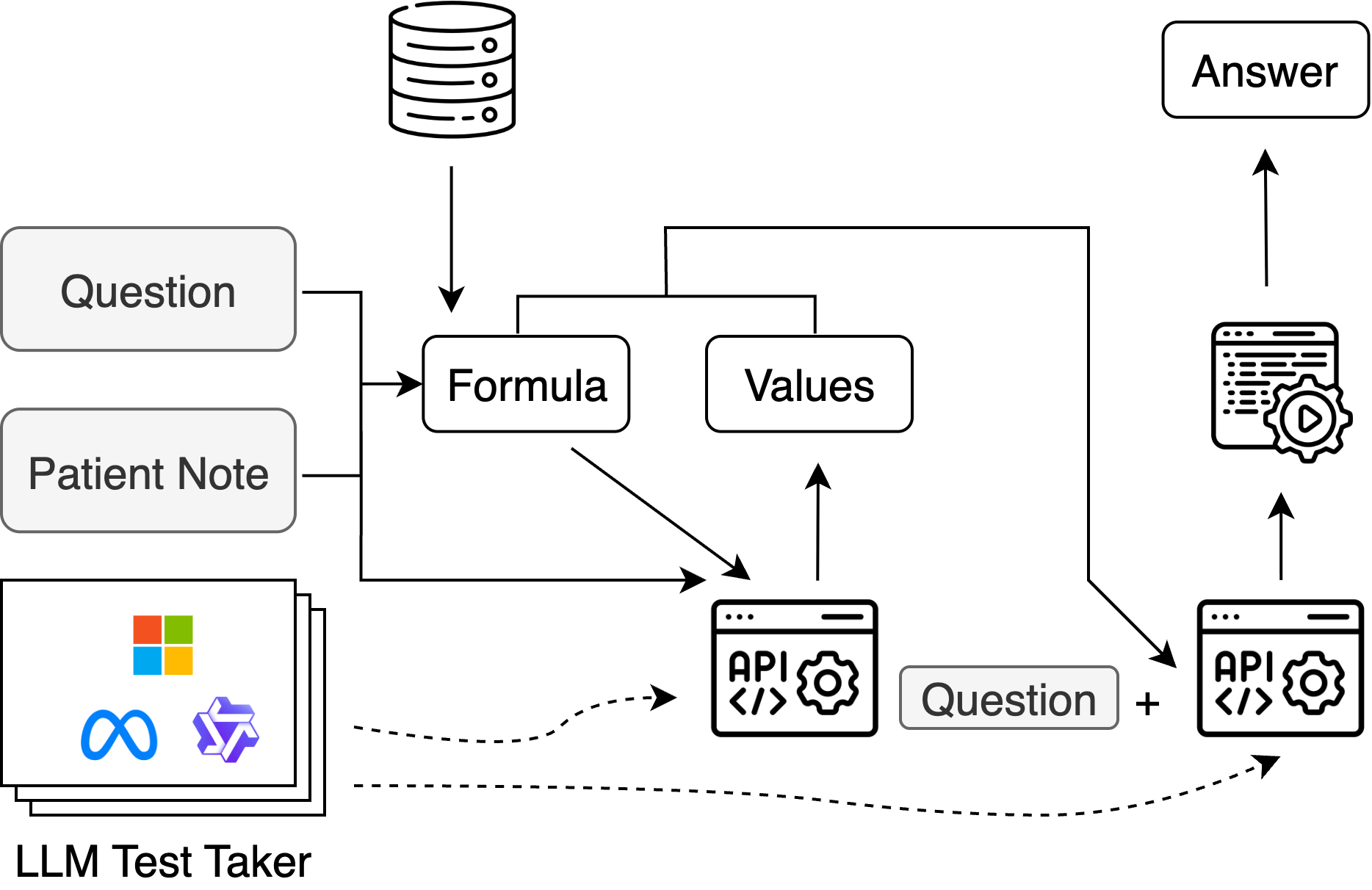}
  \caption{MedRaC Pipeline}
  \label{fig:medrac}
\end{figure}

Guided by the diagnostic insights from the step-wise evaluation and error analysis, we propose MedRaC, a modular agentic pipeline (Figure~\ref{fig:medrac}) to improve LLM performance on medical calculation tasks without any additional training. MedRaC combines Formula RAG, which embeds and indexes MDCalc formulas and task-specific descriptions so that relevant formulas can be retrieved and injected into the prompt before reasoning begins, thereby addressing formula selection errors and mitigating hallucination, and Python Code Execution, where the LLM is instructed to generate Python code representing the equation and this code is executed to produce the final result, eliminating arithmetic and rounding errors. MedRaC is designed to be plug-and-play, requiring no model fine-tuning and allowing it to be layered on top of existing LLM inference APIs. Each component targets a specific error type identified in our earlier analysis, enabling explainable and modular improvements.

\begin{table*}[t]
  \centering
  \resizebox{\linewidth}{!}{
  \begin{tabular}{lcccccccccccccc}
    \toprule
    \textbf{Model} 
    & \multicolumn{2}{c}{Direct} 
    & \multicolumn{2}{c}{CoT} 
    & \multicolumn{2}{c}{One-shot} 
    & \multicolumn{2}{c}{Self-Refine}  
    & \multicolumn{2}{c}{Medprompt} 
    & \multicolumn{2}{c}{MedRaC} \\
    \cmidrule(r){2-3} \cmidrule(r){4-5} \cmidrule(r){6-7} \cmidrule(r){8-9} \cmidrule(r){10-11} \cmidrule(r){12-13} 
    & Rule & Calc & Rule & Calc & Rule & Calc & Rule & Calc & Rule & Calc & Rule & Calc \\
    \midrule
    Phi-4-mini     & 16.52 & 2.16 & 5.01 & 2.16 & 12.09 & 16.47 & 2.06 & 6.16 & 3.24 & 10.32 & \textbf{35.10} & \textbf{68.39} \\
    LLaMA3.2-3B     & 12.39 & 0.83 & 1.47 & 3.99 & \textbf{14.75} & \textbf{18.47} & 0.88 & 2.83 & 2.95 & 16.31 & -\tablefootnote{The model fails to output executable code}& - \\
    Qwen3-4B       & 23.30 & 26.79 & 9.73 & 25.79 & \textbf{49.85} & 59.57 & 9.44 & 26.29 & 10.32 & 45.92 & 45.72 & \textbf{68.72} \\
    Qwen3-8B         & 29.20 & 42.26 & 16.52 & 38.10 & \textbf{58.70} & 62.90 & 19.76 & 40.10 & 15.93 & 53.74 & 46.61 & \textbf{74.54} \\
    LLaMA3.1-8B     & 19.17 & 2.50 & 6.78 & 8.32 & 25.07 & 20.97 & 5.90 & 6.82 & 11.21 & 28.45 & \textbf{31.27} & \textbf{70.22} \\
    Qwen3-14B       & 39.53 & 46.59 & 26.55 & 43.43 & \textbf{60.77} & 67.05 & 27.14 & 47.25 & 8.55 & 22.30 & 50.44 & \textbf{78.37} \\
        
    GPT-4o-mini     & 22.42 & 7.99 & 23.89 & 34.61 & \textbf{52.80} & 49.58 & 26.55 & 33.94 & 11.80 & 42.43 & 50.44 & \textbf{72.71} \\
    GPT-4o          & 24.48 & 13.98 & 43.07 & 43.93 & \textbf{62.24} & 54.24 & 44.84 & 44.93 & 25.96 & 56.91 & 51.03 & \textbf{64.39} \\

    \bottomrule
  \end{tabular}
  }
  \caption{Performance comparison across models and prompting strategies using LLM-aided automatic evaluation. Accuracy is reported under both rule-based and calculation-based metrics.}
  \label{tab:main-table}
\end{table*}

\section{Experiments}

We conduct all experiments on MedCalc-Bench, a benchmark comprising 1,048 physician-curated clinical calculation cases.  
Because the original release contains several obsolete or internally inconsistent records, we manually reviewed the data and had a board-certified clinician re-audit every questionable item.  
After filtering out 108 faulty or deprecated entries, we retained 940 valid cases for evaluation.  
A detailed list of the removed items, along with the rationale for each exclusion, is provided in the Appendix~\ref{sec:removed-data}.

Our primary metric is the Step-wise LLM Evaluation proposed in Section~\ref{Step-wise Evaluation}, which separately grades formula selection, entity extraction, and arithmetic computation.  

Following the benchmark guidelines, we treat zero-shot Chain-of-Thought (CoT) prompting as the main baseline. In addition to the “direct” setting, where models output only the final numerical answer, we evaluate four reasoning-oriented variants. In CoT, the model produces a detailed chain of thought along with the final answer. One-shot uses the same output format but augments the prompt with a single worked example based on the same calculator as the test case. MedPrompt implements the k-nearest-neighbor retrieval component of MedPrompt with k=3 \cite{nori2023medprompt}, without option-ordering heuristics since calculation tasks lack a multiple-choice structure. Finally, Self-Refine asks the model to critique its own response and revise the solution if an error is detected, terminating early when no error is reported, with at most five refinement rounds.

\subsection{Evaluation Results}
Table~\ref{tab:main-table} summarizes performance across a diverse suite of closed- and open-source LLMs of varying sizes.  
For the direct setting, we score only the final answer, whereas all reasoning-based variants are assessed with the automatic step-wise rubric described above.

Our MedRaC method outperforms One-shot prompting across most settings. For equation-based questions, the improvement is substantial regardless of model size, confirming the benefit of external formula retrieval and modular reasoning. For \textit{rule-based questions}, the performance gains are more nuanced. Smaller models (e.g., Phi-4-mini, LLaMA series) benefit significantly from our method over One-shot, whereas stronger models (e.g., Qwen-3, GPT-3.5) show marginal improvements or even slightly worse results.
We hypothesize two reasons for this pattern: (1) Larger models possess richer internal medical knowledge and are less reliant on external formulas, reducing the added value of MedRaC for rule-based cases. (2) The One-shot examples include not only scoring rules but also a worked-out example mapping patient notes to scores, which involves clinical reasoning. Stronger models are more capable of extracting and generalizing such implicit knowledge, enabling better transfer to new inputs.

\subsection{Validation of LLM-aided Evaluation}
We validate our evaluation pipeline from two perspectives: its ability to more effectively identify reasoning errors, and its high agreement with expert human annotations, demonstrating both improved diagnostic capability and alignment with clinical judgment.

\paragraph{Improved Detection of Reasoning Failures.}
The original MedCalc-Bench evaluates only the final numeric answer and allows a wide tolerance margin, often obscuring hallucinations or logical errors in intermediate steps. In contrast, our pipeline evaluates each stage, formula selection, variable extraction, arithmetic computation, and final answer formatting independently. This granular evaluation enables the detection of clinically significant errors that would be overlooked under final-answer-only metrics. Appendix~\ref{sec:case_study_sodium} presents a case study where an LLM generated the correct final value but introduced multiple hallucinations during intermediate reasoning; our system successfully identified these inconsistencies.

\paragraph{Alignment with Expert Annotations.}
To evaluate the reliability of our step-wise evaluation pipeline, we compare its outputs against human annotations. We randomly sampled 46 clinical calculation questions across five calculators, spanning both rule-based and equation-based tasks. Each step in our pipeline was independently annotated for correctness by both expert and non-expert evaluators.

We assessed the alignment between our evaluation pipeline and human judgments by computing pairwise agreement scores. Specifically, we measured agreement among all human annotator pairs, as well as between our error analysis pipeline and expert annotators. Agreement is defined as simple percent agreement:
\[
\text{Agreement}(a, b) = \frac{1}{n} \sum_{i=1}^{n} \mathbb{1}[a_i = b_i],
\]
where \( a_i \) and \( b_i \) are binary correctness labels from two sources (e.g., expert and pipeline), and \( \mathbb{1}[\cdot] \) is the indicator function.

Our results in Table~\ref{tab:annotator-agreement} show that LLM-based error analysis achieves higher agreement with expert annotators than non-experts, and outperforms all human annotator pairs except on the extraction task. We attribute this to the LLM's careful, step-by-step consistency in evaluating responses. These findings support the validity of our evaluation pipeline in better reflecting expert clinical judgment.

\begin{table}[h]
  \centering
  \resizebox{\linewidth}{!}{
  \begin{tabular}{lcccc}
    \toprule
    \textbf{Agreement Type} & \textbf{Formula} & \textbf{Extraction} & \textbf{Calculation} & \textbf{Answer} \\
    \midrule
    Expert--Expert      & 84.8\% & 84.8\% & 89.1\% & 95.7\% \\
    Expert--Non-Expert  & 72.3\% & 78.1\% & 66.6\% & 91.3\% \\
    LLM--Expert         & 90.2\% & 78.3\% & 88.1\% & 97.8\% \\
    All Pairs (Overall) & 77.2\% & 81.9\% & 75.7\% & 92.5\% \\
    \bottomrule
  \end{tabular}
  }
  \caption{Agreement scores (\%) across evaluation stages.}
  \label{tab:annotator-agreement}
\end{table}

\subsection{Error Type Experiments}

We compare the error type annotations produced by our LLM-based pipeline with those from human evaluators, as detailed in Appendix~\ref{section:human_annotation}, using the same experimental setup. Each annotator was asked to label all applicable error types in LLM-generated answers, and agreement was computed using Jaccard similarity:
\[
\text{Agreement}(A, B) = \frac{|A \cap B|}{|A \cup B|}
\]
where $A$ and $B$ are the sets of error types identified by two annotators.

Table~\ref{tab:annotator_llm_agreement} shows the average agreement between the LLM Judge and experts, as well as between experts and non-experts. While LLM–expert agreement is not consistently higher than human–human agreement, we observe that the LLM is reasonably aligned with expert decisions, particularly on well-defined tasks such as variable extraction and missing inputs.

These results reflect the inherent difficulty of multi-label error attribution: humans often converge on the most salient error, while LLMs evaluate each category independently and systematically. Although imperfect, the LLM’s error analysis is structured, reproducible, and offers a valuable reference point for reviewing model failures.

\begin{table}
\centering
\resizebox{\linewidth}{!}{
\begin{tabular}{lcc}
\toprule
\textbf{Error Type} & \textbf{LLM} & \textbf{Non-Expert} \\
\midrule
Arithmetic                  & 73.9\% & 95.7\% \\
Clinical Misinterpretation  & 76.1\% & 87.0\% \\
Formula                     & 73.4\% & 89.1\% \\
Variable Extraction         & 75.0\% & 76.1\% \\
Missing Variables           & 90.2\% & 100.0\% \\
Precision Errors            & 79.4\% & 93.5\% \\
\bottomrule
\end{tabular}
}
\caption{Average agreement (\%) between expert--LLM and expert--non-expert across error types.}
\label{tab:annotator_llm_agreement}
\end{table}

\paragraph{Error‐type comparison.}
We evaluated differences in output error types between Zero-Shot and MedRaC across four models, as detailed in Appendix~\ref{error-type-appendix}. Figure~\ref{fig:error_types_bar}, using the Llama3.1-8B-Instruct model as an example, illustrates that the proposed MedRaC pipeline substantially reduces the main error types relative to the Zero-Shot CoT baseline. Almost all error categories showed decreases; among them, the steepest drops were in Formula Misselection/Hallucination (\(-587\), \(-77.5\%\)), Arithmetic Error (\(-352\), \(-82.6\%\)), and Demographic Adjustment Failure (\(-105\), \(-70.9\%\)). These decreases can be understood to stem from using grounded formulas and precise programmatic calculation.
A slight increase in the low-frequency Rounding/Precision Error category likely reflects our stricter evaluation tolerance rather than a true decline in numerical performance.
We also provide the error analysis of other methods evaluated with LLaMA3.1-8B in Table~\ref{fig:error_types_bar}. Oneshot reduces many errors because curated examples guide step-by-step reasoning, though it cannot fix arithmetic mistakes since examples do not improve raw computation. Self-Refine performs better in math-heavy categories by iteratively correcting outputs, directly addressing numerical slips. In contrast, MedPrompt often underperforms Oneshot because noisy retrieved examples dilute key signals.

\begin{figure}[h]
  \includegraphics[width=\columnwidth]{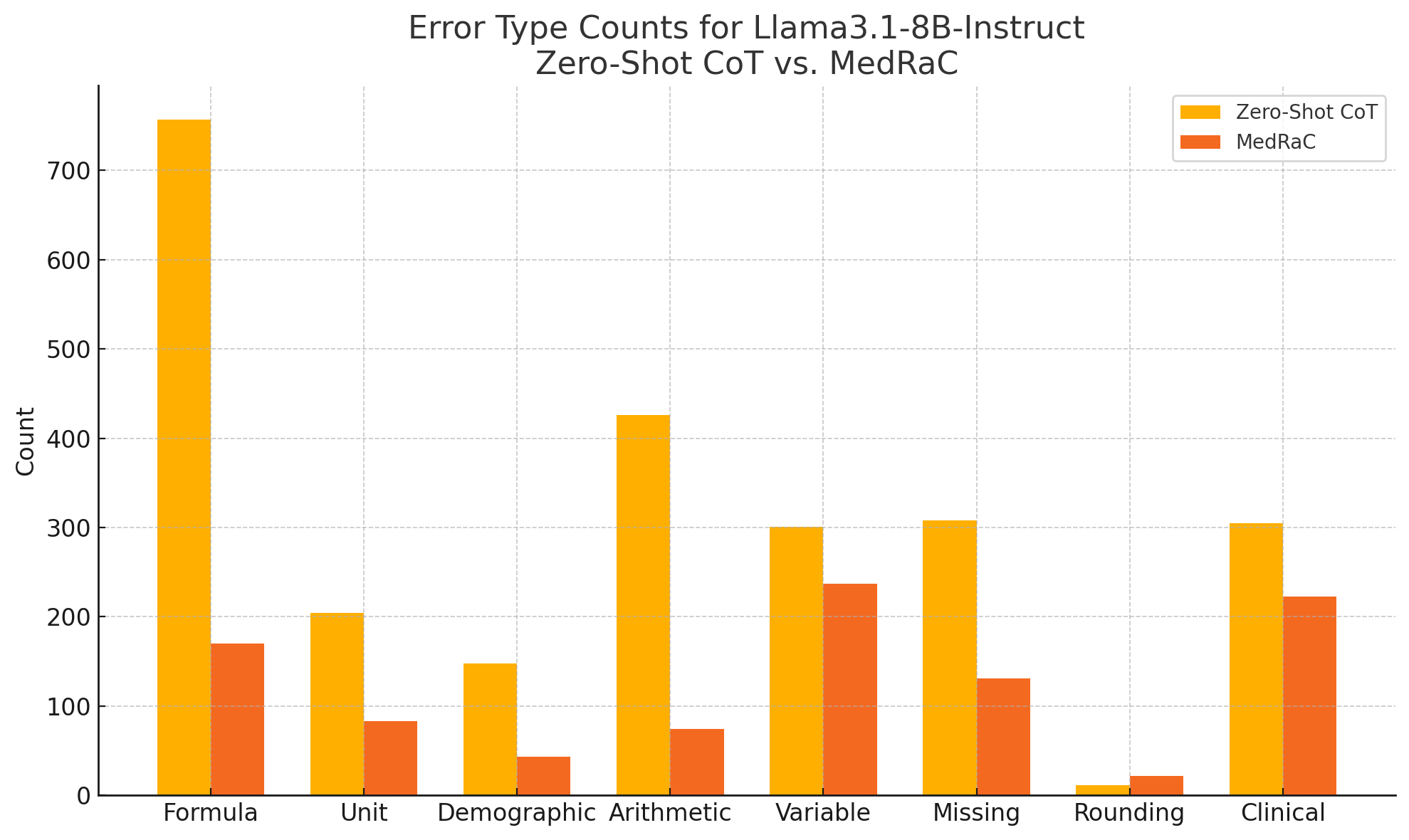}
  \caption{Error Type Counts for Llama3.1-8B-Instruct}
  \label{fig:error_types_bar}
\end{figure}

\begin{table*}[t]
\centering
\resizebox{\linewidth}{!}{
\begin{tabular}{lcccccccc}
\toprule
\textbf{Method} &
\vtop{\hbox{\strut\textbf{Formula}}\hbox{\strut\textbf{Error}}} &
\vtop{\hbox{\strut\textbf{Missing}}\hbox{\strut\textbf{Variables}}} &
\vtop{\hbox{\strut\textbf{Missing/}}\hbox{\strut\textbf{Misused}}
      \hbox{\strut\textbf{Demographic}}\hbox{\strut\textbf{Coeff.}}} &
\vtop{\hbox{\strut\textbf{Unit}}\hbox{\strut\textbf{Conversion}}\hbox{\strut\textbf{Error}}} &
\vtop{\hbox{\strut\textbf{Arithmetic}}\hbox{\strut\textbf{Errors}}} &
\vtop{\hbox{\strut\textbf{Rounding/}}\hbox{\strut\textbf{Precision}}} &
\vtop{\hbox{\strut\textbf{Incorrect}}\hbox{\strut\textbf{Variable}}\hbox{\strut\textbf{Extraction}}} &
\vtop{\hbox{\strut\textbf{Clinical}}\hbox{\strut\textbf{Mis-}}\hbox{\strut\textbf{interpretation}}} \\
\midrule
CoT         & 757 & 308 & 148 & 204 & 426 & 11 & 301 & 305 \\
Oneshot     & 295 & 152 & 44  & 82  & 455 & 5  & 194 & 241 \\
Self-Refine & 818 & 716 & 29  & 46  & 101 & 5  & 370 & 118 \\
Medprompt   & 477 & 271 & 60  & 97  & 430 & 54 & 307 & 295 \\
MedRaC      & 170 & 131 & 43  & 83  & 74  & 22 & 237 & 223 \\
code-only   & 776 & 430 & 93  & 181 & 91  & 17 & 416 & 318 \\
rag-only    & 211 & 142 & 44  & 107 & 318 & 35 & 213 & 238 \\
\bottomrule
\end{tabular}
}
\caption{Error counts by method and error type.}
\label{tab:error_counts}
\end{table*}

\paragraph{Attribution of gains.}
Formula retrieval provides the model with the exact equation and relevant demographic terms, which reduces hallucinations and incorrect formulations, thereby lowering formula-related and adjustment errors. Code execution delegates arithmetic operations to Python, preventing mistakes such as incorrect operation order or unit miscalculations and yielding roughly an 83\% reduction in arithmetic errors along with fewer unit conversion issues. An ablation study demonstrating the individual contributions of these two techniques will be presented in Section~\ref{section:ablation}.

\paragraph{Residual challenges.}
Error types that depend on nuanced clinical understanding, such as Incorrect Variable Extraction ($-64$, $-21.3\%$) and Clinical Misinterpretation ($-82$, $-26.9\%$), show relatively limited improvement. These cases often require background medical knowledge or familiarity with clinical reasoning, which current LLMs lack. Even when a correct formula is available, models may struggle if the necessary medical context is implicit or not explicitly encoded. This indicates that upstream information extraction remains a bottleneck.

\section{Ablation Studies}
\label{section:ablation}

To analyze the contribution of each MedRaC component, we conduct controlled ablations.

\paragraph{Formula RAG}

We compare MedRaC with and without retrieval, keeping the rest of the pipeline fixed. In the no-retrieval variant, the LLM is prompted to infer relevant background information before extracting the value. As shown in Table~\ref{tab:rag-ablation}, accuracy drops sharply from 64.68\% to 25.64\% without retrieval. The formula stage becomes the primary failure point, with its First Error Attribution Rate (FE) rising to 71.96\% and its Conditional Correctness (CC) falling to 7.34\%. These results suggest that retrieval is crucial for selecting accurate formulas and for subsequent reasoning.

\begin{table}[ht]
  \centering
  \resizebox{\linewidth}{!}{
  \begin{tabular}{lcc}
    \hline
    \textbf{Components} & MedRaC & MedRaC w/o RAG \\
    \hline
    Acc \% $\uparrow$  & 64.68 & 25.64 \\
    \hline
    Formula FE \% $\downarrow$   &20.78 & 71.96 \\
    \hline
    Formula CC \%  $\uparrow$ & 92.66 & 46.49 \\
    \hline
  \end{tabular}
  }
  \caption{Comparing accuracies w/ and w/o RAG}
  \label{tab:rag-ablation}
\end{table}

\paragraph{Code}

We compare variants with and without code execution. In place of code generation, the model is asked to produce chain-of-thought reasoning to calculate the requested value. Since DeepSeek-v3 struggles to accurately assess Python code, we rely on reasoning models as judges; results for DeepSeek-v3 are shown here, with the rest reported in Appendix~\ref{section:ablation}. Across different judge models, the inclusion of the Code component consistently reduces error rates.

\begin{table}[ht]
  \centering
  \resizebox{\linewidth}{!}{
  \begin{tabular}{lcc}
    \hline
    \textbf{Components} & MedRaC & MedRaC w/o Code \\
    \hline
    Acc \% $\uparrow$ & 64.68 & 53.09 \\
    \hline
    Calc FE \% $\downarrow$   &3.23 & 31.88 \\
    \hline
    Calc CC \%  $\uparrow$ & 97.82 & 76.52 \\
    \hline
  \end{tabular}
  }
  \caption{Comparing accuracies w/ and w/o Code}
  \label{tab:code-ablation}
\end{table}

\paragraph{Memory Scaling} We expand the formula bank from 55 to 785 formulas and evaluate retrieval performance using OpenAI’s embedding models. The smaller set corresponds exactly to the 55 calculators in MedCalc-Bench, while the larger set includes nearly all formulas from the MDCalc website’s evidence sections.\footnote{Formulas were obtained via web scraping. Ensure proper licensing and data usage compliance.}

Retrieval is considered successful if any of the top-$k$ retrieved formulas match the ground-truth formula for a given question. As shown in Table~\ref{tab:memory_scaling}:
\begin{itemize}
    \item All three embedding models achieve 100\% top-2 accuracy in both the 55- and 785-formula settings.
    \item Even with a 14$\times$ increase in formula count, top-1 accuracy remains high. For instance, text-embedding-ada-002 maintains 100\%, while text-embedding-3-large and text-embedding-3-small still achieve over 96\%.
\end{itemize}

These results suggest that retrieval-augmented methods are especially well-suited for medical calculation tasks, not because of any inherent superiority of RAG itself, but due to the unique nature of medical formulas. These formulas are highly structured, semantically distinct, and domain-specific, which allows embedding-based retrieval to remain robust even as the size of the knowledge base increases. This makes our method practically scalable to a much broader range of clinical calculators beyond those in MedCalc-Bench.

\section{Discussion and Conclusion}

Medical calculations are not just a numeric task. They represent structured, high-stakes reasoning in clinical workflows. Our study reveals that existing evaluation metrics, which focus solely on final answer accuracy, often fail to capture critical reasoning failures such as formula misuse, variable misinterpretation, or arithmetic errors. These oversights may result in overly optimistic assessments of model safety and applicability.

We introduce a stepwise evaluation framework and a structured error taxonomy that enable more transparent, diagnostic, and actionable feedback on model behavior. Furthermore, our MedRaC pipeline improves performance without additional training by augmenting model reasoning with explicit retrieval and executable code. Through controlled ablations and human-aligned validation, we show that each component directly mitigates failure modes in clinical computation.

Importantly, our findings point to a broader methodological shift: as language models are deployed in safety-critical domains, evaluating intermediate reasoning and domain-grounded correctness becomes essential. This work advocates for domain-aware, explanation-oriented evaluation practices that bridge the gap between model development and real-world deployment. By prioritizing interpretability and modular error analysis over end-task scores, we take a step toward safer, more trustworthy AI systems that serve beyond NLP’s traditional boundaries.

\section{Limitations}
While our step-wise evaluation framework and MedRaC pipeline provide more granular insight into LLM reasoning in medical calculations, several limitations remain.
First, our benchmark currently focuses on structured, single-turn tasks involving well-defined formulas. This setup may not accurately capture the ambiguity, context switching, or exception handling that are common in real-world clinical reasoning.

Second, although our dataset covers 55 diverse calculators and our RAG component scales to hundreds more, all experiments were conducted in English, using curated clinical notes. The generalizability of our results to multilingual settings, noisy EHR data, or patient-facing dialogue remains to be studied.

Third, the correctness judgments at each reasoning step rely on LLM-as-Judge evaluation. While we validated this against expert annotations, LLM-based evaluation may still be prone to error propagation, especially for subtle clinical misinterpretations.

Finally, while MedRaC enhances factual reliability through modular design, it assumes access to accurate formula banks and structured variables. Future work should explore more open-ended clinical reasoning and integrate real-time human oversight in deployment settings.

\section{Ethics Statement}
This study uses only publicly available and anonymized clinical data, including physician-written vignettes and structured case reports from sources such as PubMed Central. No identifiable patient data were accessed or used.

The MedCalc-Bench dataset and the associated MedRaC framework are designed solely for evaluating and improving LLM capabilities in medical calculations under controlled conditions. They are not intended for diagnostic use or direct clinical deployment. All outputs must be reviewed and interpreted by licensed healthcare professionals.

To validate our evaluation pipeline, we engaged two medical experts in the United States to annotate reasoning steps and provide structured feedback on LLM outputs. They were compensated for their time at a rate of \$40 per hour, following academic ethical standards.

Finally, given the high-stakes nature of clinical MedCalc-Bench disqualifies a model for any real-world medical calculation task, while passing should be considered a necessary but not sufficient condition for use.
We caution against deploying LLMs for clinical decision-making without rigorous benchmarking, error attribution, and domain-specific oversight. Our work is intended to contribute to the responsible development of AI for healthcare, rather than replacing expert judgment.

\section*{Acknowledgments}

This material is the result of work supported with resources and the use of facilities at the Center for Healthcare Organization and Implementation Research, VA Bedford Health Care.

\newpage

\bibliography{custom}

@article{app11146421,
  title={What disease does this patient have? a large-scale open domain question answering dataset from medical exams},
  author={Jin, Di and Pan, Eileen and Oufattole, Nassim and Weng, Wei-Hung and Fang, Hanyi and Szolovits, Peter},
  journal={Applied Sciences},
  volume={11},
  number={14},
  pages={6421},
  year={2021},
  publisher={MDPI}
}

@inproceedings{jin-etal-2019-pubmedqa,
  title={Pubmedqa: A dataset for biomedical research question answering},
  author={Jin, Qiao and Dhingra, Bhuwan and Liu, Zhengping and Cohen, William and Lu, Xinghua},
  booktitle={Proceedings of the 2019 conference on empirical methods in natural language processing and the 9th international joint conference on natural language processing (EMNLP-IJCNLP)},
  pages={2567--2577},
  year={2019}
}

@inproceedings{pmlr-v174-pal22a,
  title={Medmcqa: A large-scale multi-subject multi-choice dataset for medical domain question answering},
  author={Pal, Ankit and Umapathi, Logesh Kumar and Sankarasubbu, Malaikannan},
  booktitle={Conference on health, inference, and learning},
  pages={248--260},
  year={2022},
  organization={PMLR}
}

@article{NEURIPS2024_99e81750,
  title={Medcalc-bench: Evaluating large language models for medical calculations},
  author={Khandekar, Nikhil and Jin, Qiao and Xiong, Guangzhi and Dunn, Soren and Applebaum, Serina and Anwar, Zain and Sarfo-Gyamfi, Maame and Safranek, Conrad and Anwar, Abid and Zhang, Andrew and others},
  journal={Advances in Neural Information Processing Systems},
  volume={37},
  pages={84730--84745},
  year={2024}
}

@article{Thirunavukarasu2023,
  title={Large language models in medicine},
  author={Thirunavukarasu, Arun James and Ting, Darren Shu Jeng and Elangovan, Kabilan and Gutierrez, Laura and Tan, Ting Fang and Ting, Daniel Shu Wei},
  journal={Nature medicine},
  volume={29},
  number={8},
  pages={1930--1940},
  year={2023},
  publisher={Nature Publishing Group US New York}
}

@article{chen2022pot,
  title={Program of thoughts prompting: Disentangling computation from reasoning for numerical reasoning tasks},
  author={Chen, Wenhu and Ma, Xueguang and Wang, Xinyi and Cohen, William W},
  journal={arXiv preprint arXiv:2211.12588},
  year={2022}
}

@article{lee2025evaluating,
  title={Evaluating step-by-step reasoning traces: A survey},
  author={Lee, Jinu and Hockenmaier, Julia},
  journal={arXiv preprint arXiv:2502.12289},
  year={2025}
}

@article{wang2022selfconsistency,
  title={Self-consistency improves chain of thought reasoning in language models},
  author={Wang, Xuezhi and Wei, Jason and Schuurmans, Dale and Le, Quoc and Chi, Ed and Narang, Sharan and Chowdhery, Aakanksha and Zhou, Denny},
  journal={arXiv preprint arXiv:2203.11171},
  year={2022}
}

@article{madaan2023selfrefine,
  title={Self-refine: Iterative refinement with self-feedback},
  author={Madaan, Aman and Tandon, Niket and Gupta, Prakhar and Hallinan, Skyler and Gao, Luyu and Wiegreffe, Sarah and Alon, Uri and Dziri, Nouha and Prabhumoye, Shrimai and Yang, Yiming and others},
  journal={Advances in Neural Information Processing Systems},
  volume={36},
  pages={46534--46594},
  year={2023}
}

@article{nori2023medprompt,
  title={From medprompt to o1: Exploration of run-time strategies for medical challenge problems and beyond},
  author={Nori, Harsha and Usuyama, Naoto and King, Nicholas and McKinney, Scott Mayer and Fernandes, Xavier and Zhang, Sheng and Horvitz, Eric},
  journal={arXiv preprint arXiv:2411.03590},
  year={2024}
}

@article{zhang2025hallucination,
  title   = {Hallucination Mitigation for Retrieval-Augmented Large Language Models: A Review},
  author  = {Zhang, Wan and Zhang, Jing},
  journal = {Mathematics},
  volume  = {13},
  number  = {5},
  pages   = {856},
  year    = {2025},
  doi     = {10.3390/math13050856},
  url     = {https://www.mdpi.com/2227-7390/13/5/856}
}

@article{amugongo2025retrieval,
  title   = {Retrieval augmented generation for large language models in healthcare: A systematic review},
  author  = {Amugongo, Lameck Mbangula and Mascheroni, Pietro and Brooks, Steven and Doering, Stefan and Seidel, Jan},
  journal = {PLOS Digital Health},
  volume  = {4},
  number  = {6},
  pages   = {e0000877},
  year    = {2025},
  doi     = {10.1371/journal.pdig.0000877},
  url     = {https://journals.plos.org/digitalhealth/article?id=10.1371/journal.pdig.0000877}
}

@article{chu2025vrag,
  title={Reducing Hallucinations of Medical Multimodal Large Language Models with Visual Retrieval-Augmented Generation}, 
  author={Yun-Wei Chu and Kai Zhang and Christopher Malon and Martin Renqiang Min},
  year={2025},
  eprint={2502.15040},
  archivePrefix={arXiv},
  primaryClass={cs.CL},
  url={https://arxiv.org/abs/2502.15040}, 
}

@article{goodell2025large,
  title={Large language model agents can use tools to perform clinical calculations},
  author={Goodell, Alex J and Chu, Simon N and Rouholiman, Dara and Chu, Larry F},
  journal={npj Digital Medicine},
  volume={8},
  number={1},
  pages={163},
  year={2025},
  publisher={Nature Publishing Group UK London}
}

@inproceedings{lightman2023let,
  title={Let's verify step by step},
  author={Lightman, Hunter and Kosaraju, Vineet and Burda, Yuri and Edwards, Harrison and Baker, Bowen and Lee, Teddy and Leike, Jan and Schulman, John and Sutskever, Ilya and Cobbe, Karl},
  booktitle={The Twelfth International Conference on Learning Representations},
  year={2024}
}

@inproceedings{shaib2023summarizing,
  title={Summarizing, simplifying, and synthesizing medical evidence using GPT-3 (with varying success)},
  author={Shaib, Chantal and Li, Millicent and Joseph, Sebastian and Marshall, Iain and Li, Junyi Jessy and Wallace, Byron C},
  booktitle={Proceedings of the 61st Annual Meeting of the Association for Computational Linguistics (Volume 2: Short Papers)},
  pages={1387--1407},
  year={2023}
}

@article{cockcroft1976prediction,
  title={Prediction of creatinine clearance from serum creatinine},
  author={Cockcroft, Donald W and Gault, Henry},
  journal={Nephron},
  volume={16},
  number={1},
  pages={31--41},
  year={1976},
  publisher={S. Karger AG Basel, Switzerland}
}

@article{initiative20102010,
  title={2010 rheumatoid arthritis classification criteria},
  author={Initiative, Collaborative},
  journal={Arthritis \& Rheumatism},
  volume={62},
  number={9},
  pages={2569--2581},
  year={2010},
  publisher={Citeseer}
}

@article{gage2001validation,
  title={Validation of clinical classification schemes for predicting stroke: results from the National Registry of Atrial Fibrillation},
  author={Gage, Brian F and Waterman, Amy D and Shannon, William and Boechler, Michael and Rich, Michael W and Radford, Martha J},
  journal={Jama},
  volume={285},
  number={22},
  pages={2864--2870},
  year={2001},
  publisher={American Medical Association}
}

@article{kung2023performance,
  title={Performance of ChatGPT on USMLE: potential for AI-assisted medical education using large language models},
  author={Kung, Tiffany H and Cheatham, Morgan and Medenilla, Arielle and Sillos, Czarina and De Leon, Lorie and Elepa{\~n}o, Camille and Madriaga, Maria and Aggabao, Rimel and Diaz-Candido, Giezel and Maningo, James and others},
  journal={PLoS digital health},
  volume={2},
  number={2},
  pages={e0000198},
  year={2023},
  publisher={Public Library of Science}
}

@article{jin2024hidden,
  title={Hidden flaws behind expert-level accuracy of multimodal GPT-4 vision in medicine},
  author={Jin, Qiao and Chen, Fangyuan and Zhou, Yiliang and Xu, Ziyang and Cheung, Justin M and Chen, Robert and Summers, Ronald M and Rousseau, Justin F and Ni, Peiyun and Landsman, Marc J and others},
  journal={NPJ Digital Medicine},
  volume={7},
  number={1},
  pages={190},
  year={2024},
  publisher={Nature Publishing Group UK London}
}

@article{li2024generation,
  title={From generation to judgment: Opportunities and challenges of llm-as-a-judge},
  author={Li, Dawei and Jiang, Bohan and Huang, Liangjie and Beigi, Alimohammad and Zhao, Chengshuai and Tan, Zhen and Bhattacharjee, Amrita and Jiang, Yuxuan and Chen, Canyu and Wu, Tianhao and others},
  journal={arXiv preprint arXiv:2411.16594},
  year={2024}
}

@article{yao2024mcqg,
  title={MCQG-SRefine: Multiple Choice Question Generation and Evaluation with Iterative Self-Critique, Correction, and Comparison Feedback},
  author={Yao, Zonghai and Parashar, Aditya and Zhou, Huixue and Jang, Won Seok and Ouyang, Feiyun and Yang, Zhichao and Yu, Hong},
  journal={arXiv preprint arXiv:2410.13191},
  year={2024}
}

@article{achiam2023gpt,
  title={Gpt-4 technical report},
  author={Achiam, Josh and Adler, Steven and Agarwal, Sandhini and Ahmad, Lama and Akkaya, Ilge and Aleman, Florencia Leoni and Almeida, Diogo and Altenschmidt, Janko and Altman, Sam and Anadkat, Shyamal and others},
  journal={arXiv preprint arXiv:2303.08774},
  year={2023}
}

@article{goodman2023accuracy,
  title={Accuracy and reliability of chatbot responses to physician questions},
  author={Goodman, Rachel S and Patrinely, J Randall and Stone Jr, Cosby A and Zimmerman, Eli and Donald, Rebecca R and Chang, Sam S and Berkowitz, Sean T and Finn, Avni P and Jahangir, Eiman and Scoville, Elizabeth A and others},
  journal={JAMA network open},
  volume={6},
  number={10},
  pages={e2336483},
  year={2023}
}

@article{decker2023large,
  title={Large language model- based chatbot vs surgeon-generated informed consent documentation for common procedures},
  author={Decker, Hannah and Trang, Karen and Ramirez, Joel and Colley, Alexis and Pierce, Logan and Coleman, Melissa and Bongiovanni, Tasce and Melton, Genevieve B and Wick, Elizabeth},
  journal={JAMA network open},
  volume={6},
  number={10},
  pages={e2336997--e2336997},
  year={2023},
  publisher={American Medical Association}
}

@article{ayers2023comparing,
  title={Comparing physician and artificial intelligence chatbot responses to patient questions posted to a public social media forum},
  author={Ayers, John W and Poliak, Adam and Dredze, Mark and Leas, Eric C and Zhu, Zechariah and Kelley, Jessica B and Faix, Dennis J and Goodman, Aaron M and Longhurst, Christopher A and Hogarth, Michael and others},
  journal={JAMA internal medicine},
  volume={183},
  number={6},
  pages={589--596},
  year={2023},
  publisher={American Medical Association}
}

@article{thirunavukarasu2023trialling,
  title={Trialling a large language model (ChatGPT) in general practice with the applied knowledge test: observational study demonstrating opportunities and limitations in primary care},
  author={Thirunavukarasu, Arun James and Hassan, Refaat and Mahmood, Shathar and Sanghera, Rohan and Barzangi, Kara and El Mukashfi, Mohanned and Shah, Sachin},
  journal={JMIR Medical Education},
  volume={9},
  number={1},
  pages={e46599},
  year={2023},
  publisher={JMIR Publications Inc., Toronto, Canada}
}

@article{yang2025unveiling,
  title={Unveiling GPT-4V's hidden challenges behind high accuracy on USMLE questions: Observational Study},
  author={Yang, Zhichao and Yao, Zonghai and Tasmin, Mahbuba and Vashisht, Parth and Jang, Won Seok and Ouyang, Feiyun and Wang, Beining and McManus, David and Berlowitz, Dan and Yu, Hong},
  journal={Journal of Medical Internet Research},
  volume={27},
  pages={e65146},
  year={2025},
  publisher={JMIR Publications Toronto, Canada}
}

@article{yao2025survey,
  title={A Survey on LLM-based Multi-Agent AI Hospital},
  author={Yao, Zonghai and Yu, Hong},
  year={2025},
  publisher={OSF}
}

@article{singhal2023large,
  title={Large language models encode clinical knowledge},
  author={Singhal, Karan and Azizi, Shekoofeh and Tu, Tao and Mahdavi, S Sara and Wei, Jason and Chung, Hyung Won and Scales, Nathan and Tanwani, Ajay and Cole-Lewis, Heather and Pfohl, Stephen and others},
  journal={Nature},
  volume={620},
  number={7972},
  pages={172--180},
  year={2023},
  publisher={Nature Publishing Group}
}

@article{sun2024effectiveness,
  title={Effectiveness of ChatGPT in explaining complex medical reports to patients},
  author={Sun, Mengxuan and Reiter, Ehud and Kiltie, Anne E and Ramsay, George and Duncan, Lisa and Murchie, Peter and Adam, Rosalind},
  journal={arXiv preprint arXiv:2406.15963},
  year={2024}
}

@article{liu2023gpteval,
  title={G-eval: NLG evaluation using gpt-4 with better human alignment},
  author={Liu, Yang and Iter, Dan and Xu, Yichong and Wang, Shuohang and Xu, Ruochen and Zhu, Chenguang},
  journal={arXiv preprint arXiv:2303.16634},
  year={2023}
}

@article{gu2024survey,
  title={A survey on llm-as-a-judge},
  author={Gu, Jiawei and Jiang, Xuhui and Shi, Zhichao and Tan, Hexiang and Zhai, Xuehao and Xu, Chengjin and Li, Wei and Shen, Yinghan and Ma, Shengjie and Liu, Honghao and others},
  journal={The Innovation},
  year={2024},
  publisher={Elsevier}
}

@article{fu2023gptscore,
  title={Gptscore: Evaluate as you desire},
  author={Fu, Jinlan and Ng, See Kiong and Jiang, Zhengbao and Liu, Pengfei},
  booktitle={Proceedings of the 2024 Conference of the North American Chapter of the Association for Computational Linguistics: Human Language Technologies (Volume 1: Long Papers)},
  pages={6556--6576},
  year={2024}
}

@article{ke2023critiquellm,
  title={Critiquellm: Scaling llm-as-critic for effective and explainable evaluation of large language model generation. CoRR, abs/2311.18702. detection for generative large language models},
  author={Ke, Pei and Wen, Bosi and Feng, Zhuoer and Liu, Xiao and Lei, Xuanyu and Cheng, Jiale and Wang, Shengyuan and Zeng, Aohan and Dong, Yuxiao and Wang, Hongning and others},
  booktitle={Proceedings of the 2023 Conference on Empirical Methods in Natural Language Processing},
  pages={9004--9017},
  year={2023}
}

@inproceedings{chen2023storyer,
  title     = {StoryER: Automatic Story Evaluation via Ranking, Rating and Reasoning},
  author    = {Chen, Hong and Vo, Duc and Takamura, Hiroya and Miyao, Yusuke and Nakayama, Hideki},
  booktitle = {Proceedings of the 2022 Conference on Empirical Methods in Natural Language Processing},
  pages     = {1739--1753},
  year      = {2022},
  publisher = {Association for Computational Linguistics},
  doi       = {10.18653/v1/2022.emnlp-main.114},
  url       = {https://aclanthology.org/2022.emnlp-main.114/}
}

@article{zheng2023judging,
  title={Judging llm-as-a-judge with mt-bench and chatbot arena},
  author={Zheng, Lianmin and Chiang, Wei-Lin and Sheng, Ying and Zhuang, Siyuan and Wu, Zhanghao and Zhuang, Yonghao and Lin, Zi and Li, Zhuohan and Li, Dacheng and Xing, Eric and others},
  journal={Advances in neural information processing systems},
  volume={36},
  pages={46595--46623},
  year={2023}
}

@inproceedings{zhang2024comprehensive,
  title={A comprehensive analysis of the effectiveness of large language models as automatic dialogue evaluators},
  author={Zhang, Chen and D'Haro, Luis Fernando and Chen, Yiming and Zhang, Malu and Li, Haizhou},
  booktitle={Proceedings of the AAAI Conference on Artificial Intelligence},
  volume={38},
  number={17},
  pages={19515--19524},
  year={2024}
}

@article{kocmi2023large,
  title={Large language models are state-of-the-art evaluators of translation quality},
  author={Kocmi, Tom and Federmann, Christian},
  journal={arXiv preprint arXiv:2302.14520},
  year={2023}
}

@article{yao2024medqa,
  title={MedQA-CS: Benchmarking Large Language Models Clinical Skills Using an AI-SCE Framework},
  author={Yao, Zonghai and Zhang, Zihao and Tang, Chaolong and Bian, Xingyu and Zhao, Youxia and Yang, Zhichao and Wang, Junda and Zhou, Huixue and Jang, Won Seok and Ouyang, Feiyun and others},
  journal={arXiv preprint arXiv:2410.01553},
  year={2024}
}

@article{brake2024comparing,
  title={Comparing Two Model Designs for Clinical Note Generation; Is an LLM a Useful Evaluator of Consistency?},
  author={Brake, Nathan and Schaaf, Thomas},
  journal={arXiv preprint arXiv:2404.06503},
  year={2024}
}

@article{wang2023notechat,
  title={NoteChat: a dataset of synthetic doctor-patient conversations conditioned on clinical notes},
  author={Wang, Junda and Yao, Zonghai and Yang, Zhichao and Zhou, Huixue and Li, Rumeng and Wang, Xun and Xu, Yucheng and Yu, Hong},
  journal={arXiv preprint arXiv:2310.15959},
  year={2023}
}

@article{tran2024rare,
  title={RARE: Retrieval-Augmented Reasoning Enhancement for Large Language Models},
  author={Tran, Hieu and Yao, Zonghai and Wang, Junda and Zhang, Yifan and Yang, Zhichao and Yu, Hong},
  journal={arXiv preprint arXiv:2412.02830},
  year={2024}
}

@article{jeong2024improving,
  title={Improving medical reasoning through retrieval and self-reflection with retrieval-augmented large language models},
  author={Jeong, Minbyul and Sohn, Jiwoong and Sung, Mujeen and Kang, Jaewoo},
  journal={Bioinformatics},
  volume={40},
  number={Supplement\_1},
  pages={i119--i129},
  year={2024},
  publisher={Oxford University Press}
}

@article{arora2025healthbench,
  title={HealthBench: Evaluating Large Language Models Towards Improved Human Health},
  author={Arora, Rahul K and Wei, Jason and Hicks, Rebecca Soskin and Bowman, Preston and Qui{\~n}onero-Candela, Joaquin and Tsimpourlas, Foivos and Sharman, Michael and Shah, Meghan and Vallone, Andrea and Beutel, Alex and others},
  journal={arXiv preprint arXiv:2505.08775},
  year={2025}
}

@article{chung2025verifact,
  title={Verifact: Verifying facts in llm-generated clinical text with electronic health records},
  author={Chung, Philip and Swaminathan, Akshay and Goodell, Alex J and Kim, Yeasul and Reincke, S Momsen and Han, Lichy and Deverett, Ben and Sadeghi, Mohammad Amin and Ariss, Abdel-Badih and Ghanem, Marc and others},
  journal={arXiv preprint arXiv:2501.16672},
  year={2025}
}

@article{croxford2025automating,
  title={Automating evaluation of AI text generation in healthcare with a large language model (LLM)-as-a-judge},
  author={Croxford, Emma and Gao, Yanjun and First, Elliot and Pellegrino, Nicholas and Schnier, Miranda and Caskey, John and Oguss, Madeline and Wills, Graham and Chen, Guanhua and Dligach, Dmitriy and others},
  journal={medRxiv},
  pages={2025--04},
  year={2025},
  publisher={Cold Spring Harbor Laboratory Press}
}

@article{tu2025towards,
  title={Towards conversational diagnostic artificial intelligence},
  author={Tu, Tao and Schaekermann, Mike and Palepu, Anil and Saab, Khaled and Freyberg, Jan and Tanno, Ryutaro and Wang, Amy and Li, Brenna and Amin, Mohamed and Cheng, Yong and others},
  journal={Nature},
  pages={1--9},
  year={2025},
  publisher={Nature Publishing Group UK London}
}

@article{shen2025let,
  title={Let's Verify Math Questions Step by Step},
  author={Shen, Chengyu and Wong, Zhen Hao and He, Runming and Liang, Hao and Qiang, Meiyi and Meng, Zimo and Zhao, Zhengyang and Zeng, Bohan and Zhu, Zhengzhou and Cui, Bin and others},
  journal={arXiv preprint arXiv:2505.13903},
  year={2025}
}

@article{li2024mediq,
  title={Mediq: Question-asking llms and a benchmark for reliable interactive clinical reasoning},
  author={Li, Stella and Balachandran, Vidhisha and Feng, Shangbin and Ilgen, Jonathan and Pierson, Emma and Koh, Pang Wei W and Tsvetkov, Yulia},
  journal={Advances in Neural Information Processing Systems},
  volume={37},
  pages={28858--28888},
  year={2024}
}

@article{
mialon2023augmented,
title={Augmented Language Models: a Survey},
author={Gr{\'e}goire Mialon and Roberto Dessi and Maria Lomeli and Christoforos Nalmpantis and Ramakanth Pasunuru and Roberta Raileanu and Baptiste Roziere and Timo Schick and Jane Dwivedi-Yu and Asli Celikyilmaz and Edouard Grave and Yann LeCun and Thomas Scialom},
journal={Transactions on Machine Learning Research},
issn={2835-8856},
year={2023},
url={https://openreview.net/forum?id=jh7wH2AzKK},
note={Survey Certification}
}

@inproceedings{pmlr-v202-gao23f,
  title={Pal: Program-aided language models},
  author={Gao, Luyu and Madaan, Aman and Zhou, Shuyan and Alon, Uri and Liu, Pengfei and Yang, Yiming and Callan, Jamie and Neubig, Graham},
  booktitle={International Conference on Machine Learning},
  pages={10764--10799},
  year={2023},
  organization={PMLR}
}

@misc{MDCalcAboutUs,
  title        = {About Us - MDCalc},
  howpublished = {\url{https://www.mdcalc.com/about-us}},
  note         = {Accessed: 2025-09-19},
  year         = {2025},
  organization = {MDCalc},
}

@inproceedings{xiong2024benchmarking,
  title={Benchmarking retrieval-augmented generation for medicine},
  author={Xiong, Guangzhi and Jin, Qiao and Lu, Zhiyong and Zhang, Aidong},
  booktitle={Findings of the Association for Computational Linguistics ACL 2024},
  pages={6233--6251},
  year={2024}
}

@inproceedings{xiong2024improving,
  title     = {Improving retrieval-augmented generation in medicine with iterative follow-up questions},
  author    = {Xiong, Guangzhi and Jin, Qiao and Wang, Xiao and Zhang, Minjia and Lu, Zhiyong and Zhang, Aidong},
  booktitle = {Pacific Symposium on Biocomputing (PSB) 2025},
  pages     = {199--214},
  year      = {2025},
  organization = {World Scientific},
  url       = {https://psb.stanford.edu/psb-online/proceedings/psb25/xiong.pdf}
}

@article{nori2023can,
  title={Can generalist foundation models outcompete special-purpose tuning? case study in medicine},
  author={Nori, Harsha and Lee, Yin Tat and Zhang, Sheng and Carignan, Dean and Edgar, Richard and Fusi, Nicolo and King, Nicholas and Larson, Jonathan and Li, Yuanzhi and Liu, Weishung and others},
  journal={arXiv preprint arXiv:2311.16452},
  year={2023}
}

@article{wang2024jmlr,
  title={Jmlr: Joint medical llm and retrieval training for enhancing reasoning and professional question answering capability},
  author={Wang, Junda and Yang, Zhichao and Yao, Zonghai and Yu, Hong},
  journal={arXiv preprint arXiv:2402.17887},
  year={2024}
}

\newpage

\appendix
\section{Removed Data List and Reasons}
\label{sec:removed-data}

From the original 1\,048 examples, we excluded 108 records (10.3\%) after a double-blinded review by two expert clinicians.  
Retaining these flawed items would have distorted both step-wise and end-to-end accuracy, so they were discarded before any model-level analysis.

\bigskip
\noindent\textbf{Calculator ID 13 — \emph{Estimated Due Date} (Equation-Based).}\\
\textit{Rows removed: all.}\\
The gestational-age equation in the benchmark spreadsheet was mis-typed, yielding uniformly incorrect targets; therefore, every instance was removed.

\medskip
\noindent\textbf{Calculator ID 28 — \emph{APACHE II Score} (Rule-Based).}\\
\textit{Rows removed: all.}\\
The ground-truth rule assigned {\small+4} points when the alveolar–arterial gradient exceeded 349 mmHg, but authoritative guidelines award {\small+3} points from 350–499 mmHg and reserve {\small+4} points for $\ge\!500$ mmHg.

\medskip
\noindent\textbf{Calculator ID 3 — \emph{FeverPAIN Score for Strep Pharyngitis} (Rule-Based).}\\
\textit{Rows removed: 451 only.}\\
Wrong ground truth answer.

\medskip
\noindent\textbf{Calculator ID N/A}\\
\textit{Rows removed (45 total):}\\
\texttt{472, 803, 473, 946, 940, 804, 764, 936, 761, 798, 930, 738, 934, 938, 789, 469, 792, 948, 944, 937, 781, 941, 507, 478, 801, 945, 931, 477, 806, 929, 763, 794, 471, 932, 481, 942, 947, 943, 805, 486, 939, 768, 810, 933, 935, 468}\\
For these cases with negative answers, the data incorrectly specifies the Lower Limit and Upper Limit in reversed order—for example, Lower Limit = –4 and Upper Limit = –5—due to improper handling of negative values. As a result, the original benchmark evaluation method always yields incorrect results.

\medskip
\noindent\textbf{Calculator ID 11 — \emph{QTc Bazett Calculator} (Equation-Based).}\\
\textit{Rows removed: all.}\\
Prompt instructions required answers in \emph{seconds}, whereas the ground-truth column stored QTc in \emph{milliseconds}, causing a systematic unit mismatch. Standard QT/QTc formulae—including Bazett—are defined in seconds.

\medskip
\noindent\textbf{Calculator ID 36 — \emph{Caprini Score (2005)} (Rule-Based).}\\
\textit{Rows removed: all.}\\
The benchmark granted {\small+1} point to every female patient. The validated rule adds this point only for pregnancy or other specific obstetric conditions; indiscriminate gender scoring overestimates risk.

\section{Inference Environment}
All runs are inference-only—no training is performed.  
We use two NVIDIA RTX~A6000 GPUs for local execution of open-source models. For GPT models, we use the default settings. For all other open-source models, we set the temperature to $0.6$, top-p to $0.95$, and repetition penalty to $1.0$. The LLM Evaluation pipeline is conducted using DeepSeek-chat as the evaluator, and Error Types are checked by DeepSeek-reasoner.

\begin{table}[h]
  \centering
  \resizebox{\linewidth}{!}{
    \begin{tabular}{lccc}
      \hline
      \textbf{Model} & \textbf{\#F} & \textbf{Top-1 (\%)} & \textbf{Top-2 (\%)} \\
      \hline
      ada-002 & 55  & 100.00 & 100.00 \\
              & 785 & 100.00 & 100.00 \\
      \hline
      3-large & 55  &  98.18 & 100.00 \\
              & 785 &  96.36 & 100.00 \\
      \hline
      3-small & 55  &  94.55 & 100.00 \\
              & 785 &  98.18 & 100.00 \\
      \hline
    \end{tabular}
  }
  \caption{Top-$k$ retrieval accuracy on formula sets of size 55 and 785 using OpenAI embedding models. “\#F” denotes number of formulas.}
  \label{tab:memory_scaling}
\end{table}

\section{Human Annotation Details}
\label{section:human_annotation}

Our human annotation study was conducted in a single session with four annotators: two medically trained experts and two students from medically related fields. Annotators were provided with the LLM-generated outputs and asked to assess their correctness. For outputs judged incorrect, they were further asked to identify the corresponding error types.

We initially selected 35 samples from code-based generations and 15 from non-code generations. During review, two samples were found to have incorrect ground truth labels. These were replaced with two additional randomly selected samples from the same calculator. However, further inspection revealed additional labeling issues. After discarding all problematic entries, the final dataset contained 46 validated samples.

\newpage
\subsection{Human Annotation Guidance}

Human Evaluation Guidelines
Evaluation Steps
Thank you very much for assisting us with data annotation. The data to be evaluated is provided in a Google Sheet, where each row corresponds to a response generated by a large language model (LLM) for a medical calculation task. These responses fall into two broad categories:
- Equation-based Calculation Tasks: Questions that require explicit mathematical formulae (e.g., computing BMI).
- Rule-based Calculation Tasks: Questions that involve assigning points based on clinical criteria and summing them (e.g., Wells score).
Step 1: Answer Validation
Please read the Patient Note, and carefully read the Question and the Ground Truth Explanation. Then assess the LLM Answer.

The answer consists of four steps: formula, extracted values, calculation, and final answer. Formula and extracted values are complemented by their corresponding reasons. 
Please review each field carefully and verify that the steps and results are accurate. If it’s not fully correct, please refer to the Possible Error Types section on the next page and select all errors that are present. 
Note: ONLY if ALL intermediate steps and the final result are correct is the LLM Answer considered correct.
Note2: Formula correctness is assessed on both the formula field and the actual formula used during calculation.
Step 2: LLM Self-Evaluation Assessment
Next, review the LLM Evaluation columns. These involves another model’s assessment of previous LLM’s  performance in terms of:
- Formula correctness
- Extracted values correctness
- Mathematical calculation correctness

For each column, please judge the evaluation result as one of the following:
 Correct / Correct but explanation flawed /  Incorrect
We greatly appreciate your help and detailed annotations!
Possible Error Types
Incorrect Formula Selection
The wrong medical formula is used for the given clinical scenario.
Example: Using the Cockcroft-Gault equation to estimate GFR in an AKI patient instead of CKD-EPI.
Internal Formula Logic or Parameter Errors (Hallucination)
- Equation-based Questions:
  The overall formula structure is correct, but components such as terms, coefficients, or exponents are incorrect or hallucinated.
  Example: Writing Framingham QTc as QT + 154*1 – RR, where missing parentheses result in a miscalculation.

- Rule-based Questions:
  Scoring items are missing or fabricated.
  Example: Omitting “recent surgery” from the Wells score or adding a non-existent “family history” item.
Incorrect Variable Extraction
Values extracted from the patient note are incorrect in terms of number, timing, or unit.
Example: Extracting “heart rate = 76 bpm” as 176 bpm, or using a lab value from a previous visit instead of the current one.
Clinical Misinterpretation (Rule-based Only)
Misunderstanding the clinical implications of a symptom or finding can lead to incorrect scoring.
Example: “Abdomen was diffusely distended” suggests mild ascites (+2), but the model assumes no ascites and assigns +1.
Missing Variables
The model fails to extract the required inputs, making it impossible to complete the calculation.
Example: Missing weight or race information causes incomplete or halted computation.

Unit Conversion Errors
Units are not converted correctly before calculation, resulting in serious numerical errors.
Example: Using 134 µmol/L creatinine in the MDRD formula without converting to mg/dL.
Missing or Misused Demographic/Adjustment Coefficients
Important adjustment factors, such as gender, race, BMI-based weight corrections, or pregnancy status, are omitted or misused.
Example: Not applying a 0.85 coefficient for female patients in the Cockcroft-Gault equation.
Arithmetic Errors
- Equation-based Questions:
  Incorrect mathematical operations, such as order-of-operations errors or basic calculation mistakes.
  Example: Writing (A + B) * C as A + B * C, or calculating 3 × 3 as 10.

- Rule-based Questions:
  Correct scoring items are identified, but summed incorrectly.
  Example: Adding 1 + 2 + 1 and mistakenly writing 5.
Rounding / Precision Errors
Rounding is too aggressive or insufficient, leading to clinically significant inaccuracies.
Use the number of decimal places in the LLM’s answer to determine the required precision, up to a maximum of 2 decimal places.
If the LLM returns 10.65, evaluate it to 2 decimal places with a tolerance of ±0.005.
If it returns 10.7, use 1 decimal place with a tolerance of ±0.05.
If it returns 10.6512, use two decimal places with a tolerance of ±0.05.
NOTE: The final answer has already been pre-checked against the ground truth answer in the LLM Answer Eval column. You do not need to manually re-check it based on this precision rule.
This error type should be marked when rounding errors or insufficient precision in intermediate or final steps cause the final answer to fall outside the tolerance range/result in “Incorrect” in the Answer Evaluation Column.
\newpage

\section{Categorized Evaluation Results}
We present categorized evaluation results in Table~\ref{tab:specialty_results_stacked}.
MedRaC delivers substantial improvements in specialties that depend heavily on numerical calculations—such as Nephrology (39.7\% → 90.4\%), Thrombosis/Hematology (28.8\% → 76.3\%), Clinical Pharmacology (5.0\% → 65.0\%), and Endocrinology \& Metabolism (65.6\% → 90.2\%). These gains stem from two design features: grounding formula selection in trusted medical knowledge to reduce hallucinations, and executing code to eliminate arithmetic errors.  Although MedRaC does not outperform all baselines in every domain (e.g., Oneshot attains higher scores in Pulmonology and Hepatology), it achieves the most consistent and large-scale improvements in areas where computational fidelity is paramount.

By contrast, smaller gains are observed in domains such as General Practice/Family Medicine (10.0\% → 40.0\%) and Hepatology/Gastroenterology (16.9\% → 66.2\%), where success depends more on clinical judgment and contextual interpretation than on direct computation. A full comparison across models of varying parameter scales is included in the appendix.

Model scale further differentiates performance across domain types. In narrative-heavy, guideline-driven specialties (e.g., Hepatology/Gastroenterology, Infectious Disease, General Practice/Family Medicine), larger models within the same family (e.g., Qwen, LLaMA) exhibit stronger clinical recall and decision-making, reflecting the benefits of broader contextual reasoning. Conversely, in deterministic, calculation-intensive domains (e.g., Endocrinology \& Metabolism, Obstetrics \& Gynecology), even smaller models paired with code execution approach the performance ceiling. Beyond this point, increasing model size yields diminishing returns and may occasionally introduce over-generation or minor regressions.

\begin{table*}[t]
\centering
\resizebox{\linewidth}{!}{
\begin{tabular}{lccccc}
\toprule
\textbf{Specialty} &
\textbf{MedRaC} &
\textbf{CoT} &
\textbf{MedPrompt} &
\textbf{Self-Refine} &
\textbf{One-shot} \\
\midrule
Nephrology &
\shortstack{\textbf{198 / 219}\\\textbf{(90.41\%)}} &
\shortstack{87 / 219\\(39.73\%)} &
\shortstack{15 / 219\\(6.85\%)} &
\shortstack{97 / 219\\(44.29\%)} &
\shortstack{169 / 219\\(77.17\%)} \\
\midrule
Cardiology &
\shortstack{\textbf{166 / 236}\\\textbf{(70.34\%)}} &
\shortstack{76 / 236\\(32.20\%)} &
\shortstack{82 / 236\\(34.75\%)} &
\shortstack{103 / 236\\(43.64\%)} &
\shortstack{153 / 236\\(64.83\%)} \\
\midrule
Thrombosis/Hematology &
\shortstack{\textbf{45 / 59}\\\textbf{(76.27\%)}} &
\shortstack{17 / 59\\(28.81\%)} &
\shortstack{17 / 59\\(28.81\%)} &
\shortstack{25 / 59\\(42.37\%)} &
\shortstack{44 / 59\\(74.58\%)} \\
\midrule
Pulmonology \& Critical Care &
\shortstack{78 / 100\\(78.00\%)} &
\shortstack{40 / 100\\(40.00\%)} &
\shortstack{33 / 100\\(33.00\%)} &
\shortstack{54 / 100\\(54.00\%)} &
\shortstack{\textbf{93 / 100}\\\textbf{(93.00\%)}} \\
\midrule
Hepatology/Gastroenterology &
\shortstack{43 / 65\\(66.15\%)} &
\shortstack{11 / 65\\(16.92\%)} &
\shortstack{8 / 65\\(12.31\%)} &
\shortstack{23 / 65\\(35.38\%)} &
\shortstack{\textbf{50 / 65}\\\textbf{(76.92\%)}} \\
\midrule
Endocrinology \& Metabolism &
\shortstack{\textbf{110 / 122}\\\textbf{(90.16\%)}} &
\shortstack{80 / 122\\(65.57\%)} &
\shortstack{41 / 122\\(33.61\%)} &
\shortstack{91 / 122\\(74.59\%)} &
\shortstack{105 / 122\\(86.07\%)} \\
\midrule
Obstetrics \& Gynecology &
\shortstack{\textbf{38 / 40}\\\textbf{(95.00\%)}} &
\shortstack{36 / 40\\(90.00\%)} &
\shortstack{27 / 40\\(67.50\%)} &
\shortstack{36 / 40\\(90.00\%)} &
\shortstack{27 / 40\\(67.50\%)} \\
\midrule
Infectious Disease &
\shortstack{\textbf{26 / 39}\\\textbf{(66.67\%)}} &
\shortstack{16 / 39\\(41.03\%)} &
\shortstack{8 / 39\\(20.51\%)} &
\shortstack{10 / 39\\(25.64\%)} &
\shortstack{19 / 39\\(48.72\%)} \\
\midrule
Clinical Pharmacology &
\shortstack{\textbf{26 / 40}\\\textbf{(65.00\%)}} &
\shortstack{2 / 40\\(5.00\%)} &
\shortstack{22 / 40\\(55.00\%)} &
\shortstack{4 / 40\\(10.00\%)} &
\shortstack{22 / 40\\(55.00\%)} \\
\midrule
General Practice/Family Medicine &
\shortstack{8 / 20\\(40.00\%)} &
\shortstack{2 / 20\\(10.00\%)} &
\shortstack{1 / 20\\(5.00\%)} &
\shortstack{3 / 20\\(15.00\%)} &
\shortstack{\textbf{9 / 20}\\\textbf{(45.00\%)}} \\
\bottomrule
\end{tabular}
}
\caption{Correct counts and percentages across specialties (best per specialty in bold).}
\label{tab:specialty_results_stacked}
\end{table*}

\section{Error Type Results of other Models}

\label{error-type-appendix}

\begin{figure}[!ht]
  \includegraphics[width=0.98\columnwidth]{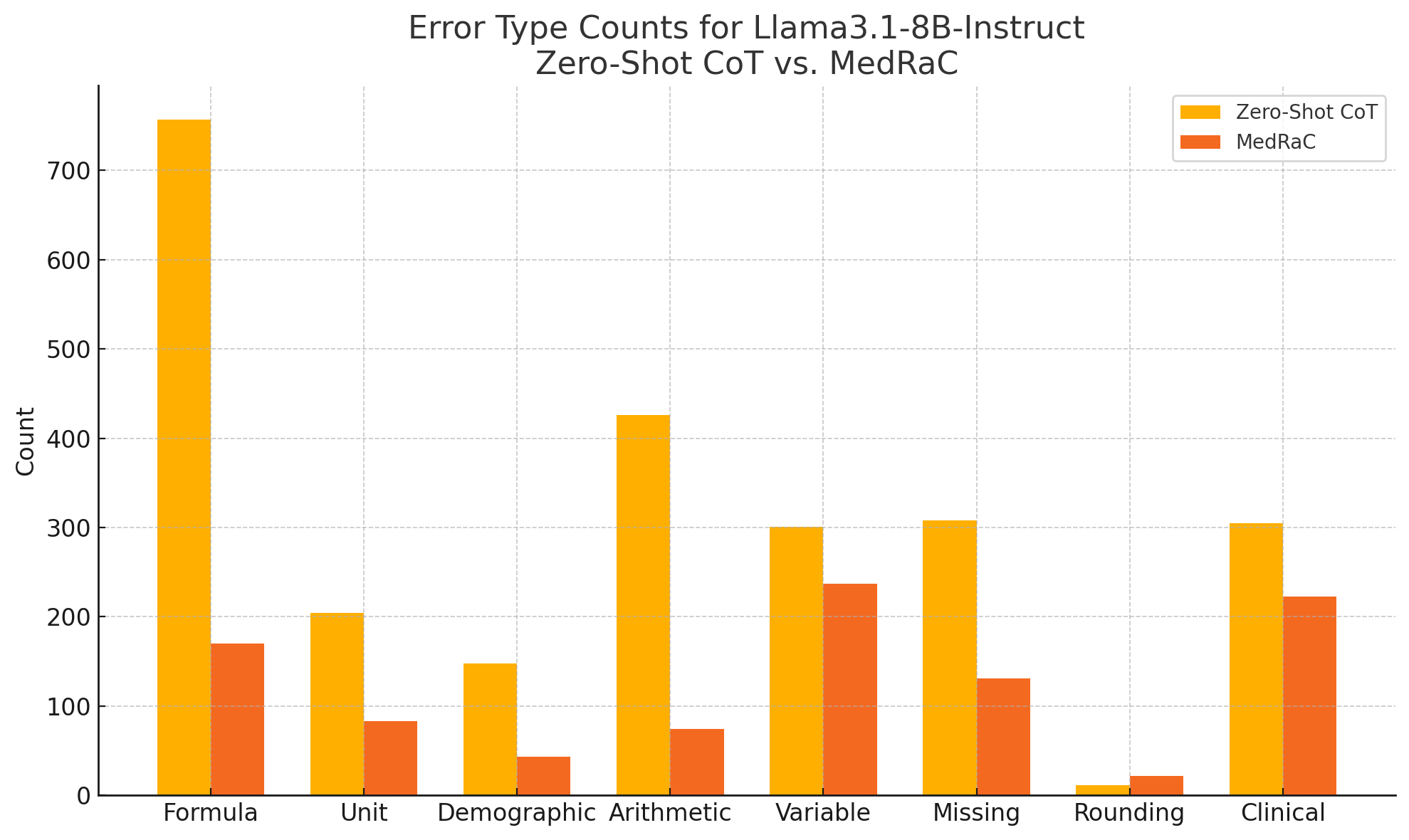}
  \captionsetup{labelformat=empty}
  \caption{Figure 4: Error Type Counts for Llama3.1-8B-Instruct}
\end{figure}

\begin{figure}[!ht]
  \includegraphics[width=0.98\columnwidth]{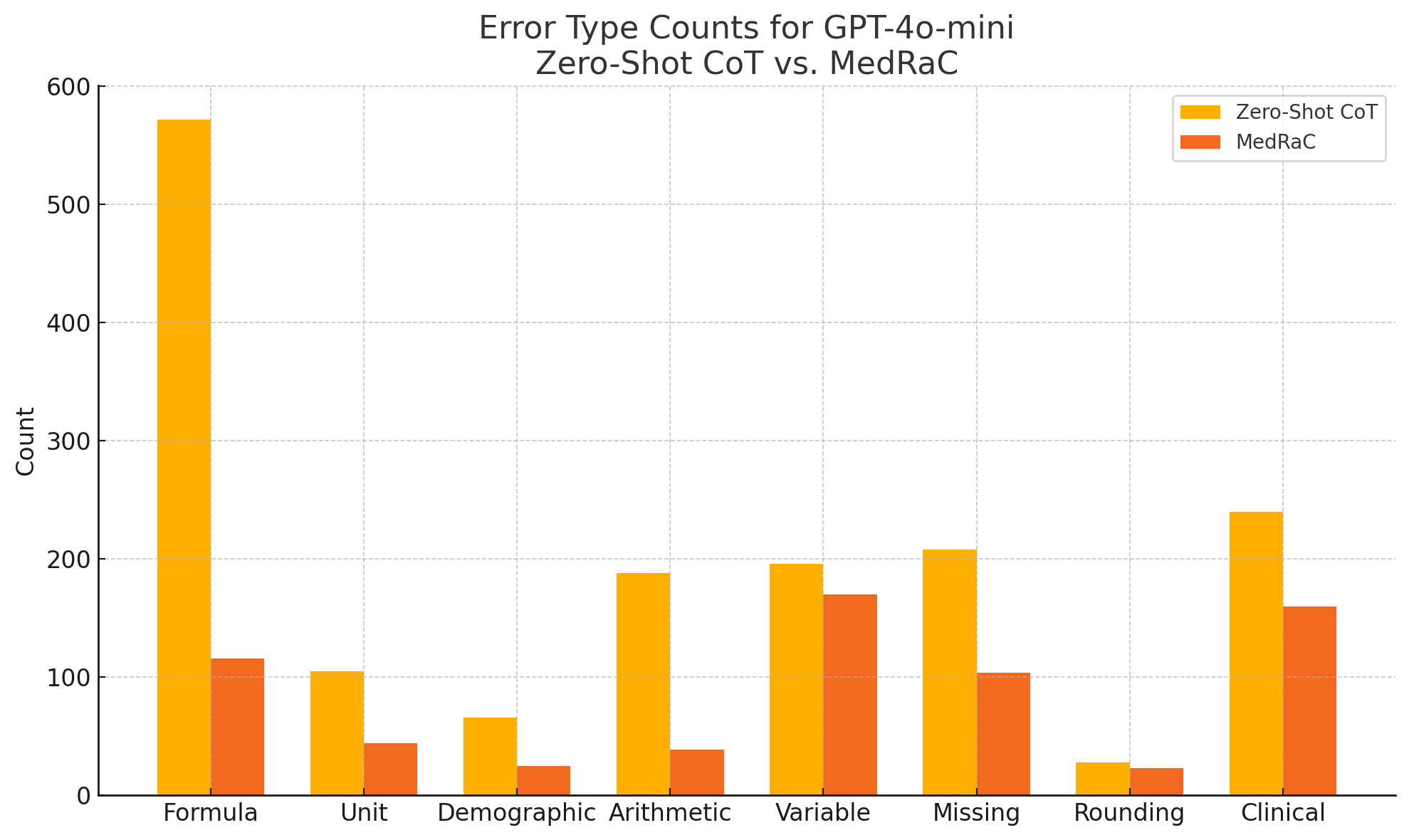}
  \caption{Error Type Counts for gpt-4o-mini}
  \label{fig:error_types_gpt}
\end{figure}

\begin{figure}[!ht]
  \includegraphics[width=0.98\columnwidth]{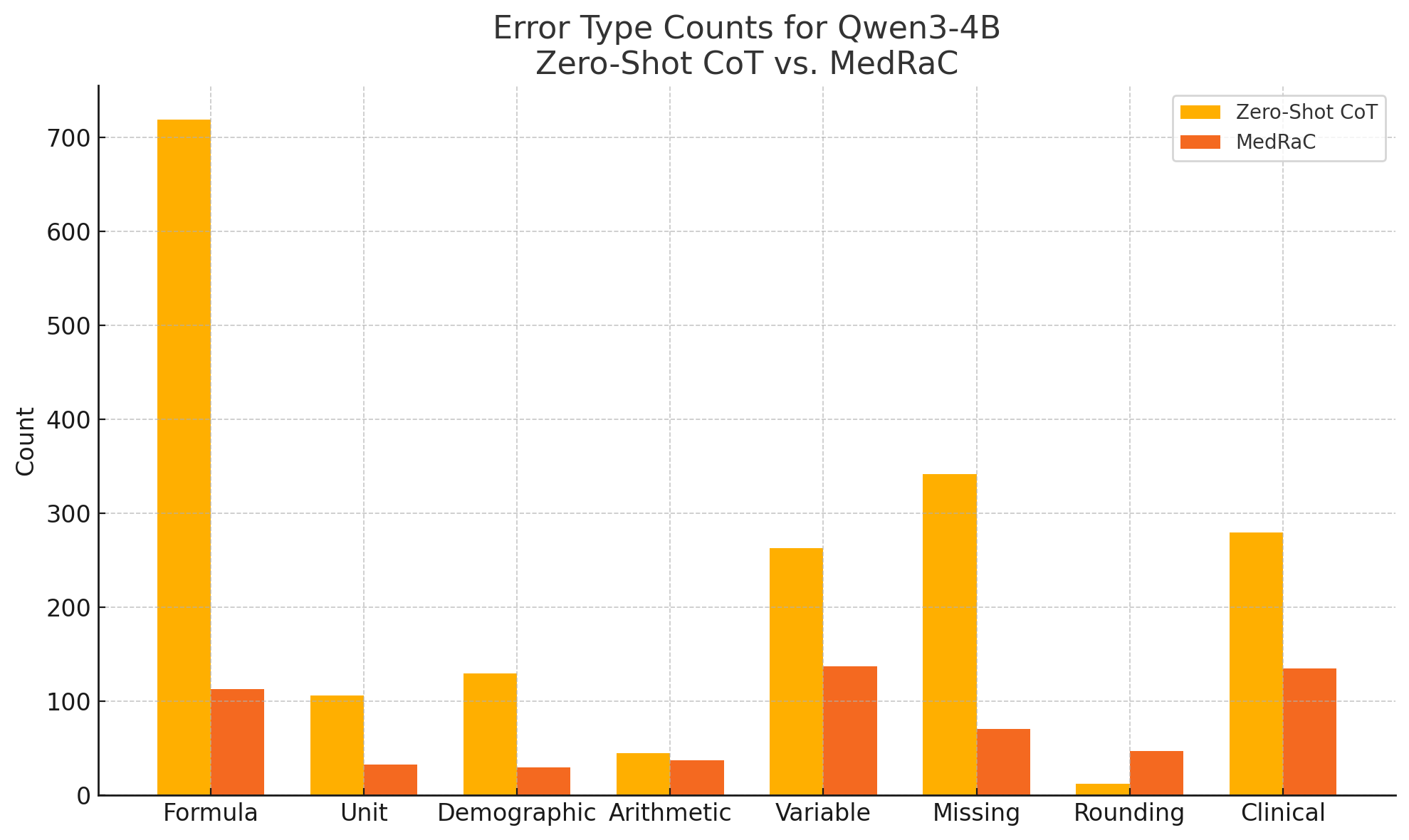}
  \caption{Error Type Counts for qwen3-4B}
  \label{fig:error_types_qwen4}
\end{figure}

\begin{figure}[!ht]
  \includegraphics[width=0.98\columnwidth]{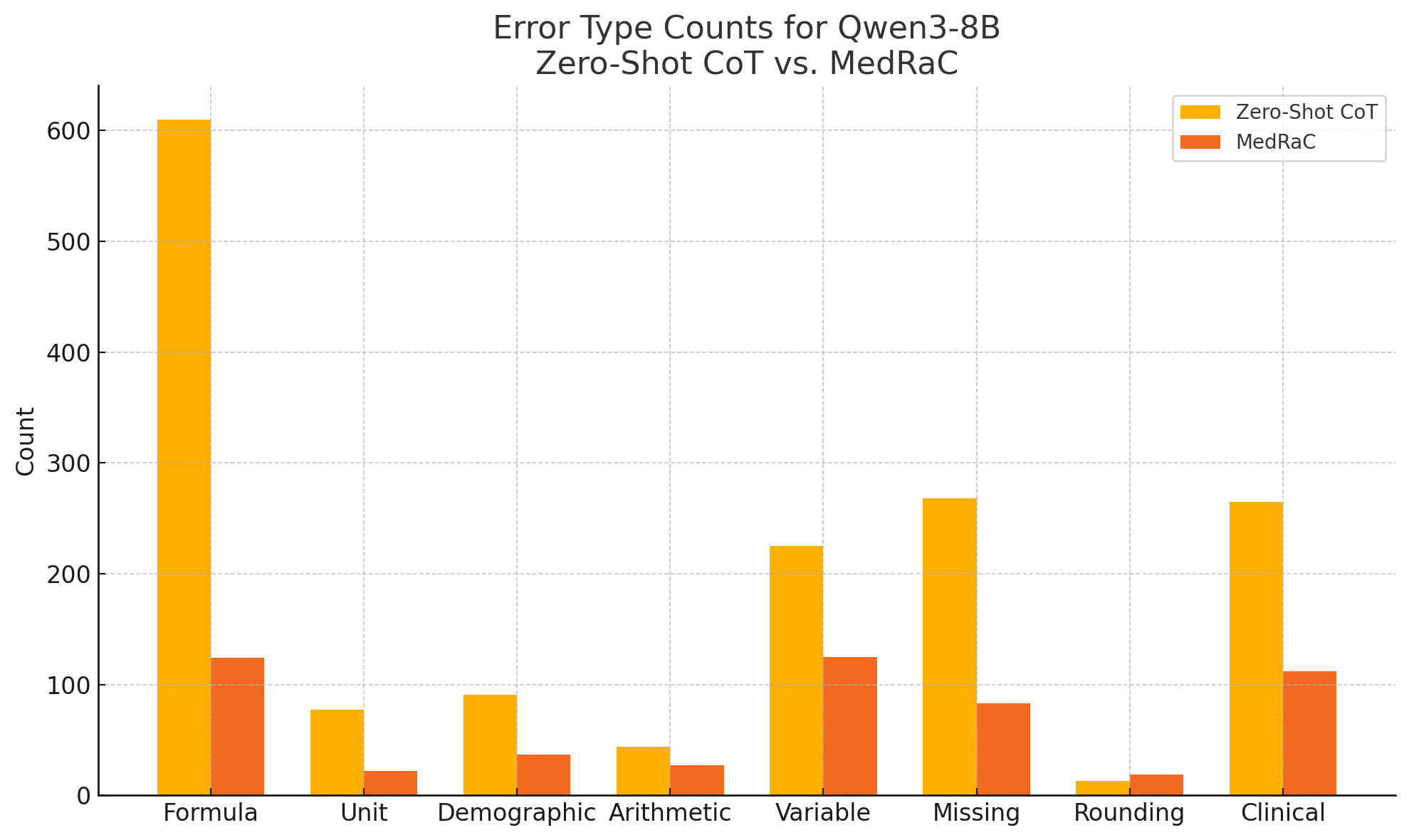}
  \caption{Error Type Counts for qwen3-8B}
  \label{fig:error_types_qwen8}
\end{figure}

Including Llama3.1-8B-Instruct, we evaluated a total of four models under both Zero-shot CoT and MedRaC: two reasoning models (GPT-4o-mini, Qwen3-4B) and two general-purpose models (Llama3.1-8B-Instruct, Qwen3-8B). The results are consistent with the trend observed for Llama3.1-8B-Instruct. In addition, we find that reasoning models show greater improvements in value extraction. 

\section{Additional Ablation Results}
\label{section:ablation}
We additionally report the evaluation of the code component by GPT-4.1 and GPT-4o-mini.
\begin{table}[h]
\centering
\resizebox{\linewidth}{!}{
\begin{tabular}{lccc}
\toprule
\textbf{Evaluation} & \textbf{MedRaC} & \textbf{w/o Code} & \\
\midrule
\multicolumn{4}{l}{\textit{Formula Error (FE) $\downarrow$}} \\
DeepSeek-R1 (reasoning) & 3.23 & 31.88 & \\
GPT-4.1                 & 2.69 & 7.02  & \\
GPT-4o-mini (reasoning) & 0.85 & 16.95 & \\
\midrule
\multicolumn{4}{l}{\textit{Calculation Correctness (CC) $\uparrow$}} \\
DeepSeek-R1 (reasoning) & 97.82 & 76.52 & \\
GPT-4.1                 & 98.97 & 96.38 & \\
GPT-4o-mini (reasoning) & 99.53 & 86.95 & \\
\bottomrule
\end{tabular}
}
\caption{Ablation results of MedRaC with and without the Code component. FE = Formula Error (lower is better), CC = Calculation Correctness (higher is better).}
\label{tab:medrac_ablation_long}
\end{table}

\onecolumn

\section{Case study: Hallucinations in LLM's answer and how our LLM Eval pipeline detects them}
\label{sec:case_study_sodium}

\subsection*{Basic record (Row 369, \textit{Sodium Correction for Hyperglycemia} calculator)}
\begin{description}[leftmargin=2.2cm,style=nextline]
  \item[Patient] 57-year-old \textbf{male}; Na\textsubscript{meas}\,$=$ 127 mmol/L (127 mEq/L); glucose\,$=$ 527 mg/dL
  \item[Clinical note] Diabetic foot with massive hyperglycaemia and haemodynamic instability (full narrative in dataset)
  \item[Question] ``What is the corrected sodium concentration (mEq/L) using the Hillier 1999 equation at admission?’’
\end{description}

\subsection*{Gold-standard reasoning}
Hillier’s formula \citeyearpar{Hillier1999}:
\[
\text{Na\textsubscript{corr}} \;=\;
\text{Na\textsubscript{meas}}
\;+\;0.024\,\bigl(\text{glucose}-100\bigr).
\]
\[
127 + 0.024\,(527-100)=\mathbf{137.248\;\text{mEq/L}}.\]
Baseline tolerance (dataset,±6.8624): \([130.39,\;144.11]\) mEq/L.

\subsection*{LLM original answer (excerpt)}
\begin{quote}\small
\emph{“Corrected sodium (mEq/L) = 127 + 0.016×527  
$=127+8.432=135.432$ mEq/L.”}
\end{quote}

\subsection*{Why the baseline benchmark says ``Correct’’}
The Medcalc benchmark inspects only whether the final number lies within the broad interval above.  
Because $135.432\in[130.39,144.11]$, the response is labelled \textbf{“Correct’’}, even though the equation is mis-specified.

\subsection*{Our stricter numeric rule}
We judge the final figure to the next decimal place beyond the model’s precision (max.\ two places).  
The LLM output has \textbf{three} decimals, so we round to two and require agreement within $\pm0.005$:
\[
\bigl|\,135.43-137.25\,\bigr| = 1.82 \;>\; 0.005,
\]
hence the answer is \textbf{incorrect} despite passing the coarse range check.

\subsection*{Step-by-step LLM Eval verdict}
\begin{center}
\begin{tabular}{@{}lcc@{}}
\toprule
\textbf{Component} & \textbf{Result} & \textbf{Key comment}\\
\midrule
Formula selection & \textcolor{red}{Incorrect} & Used $0.016\times\text{glucose}$ and omitted $-100$.\\
Entity extraction & Correct & Na\,$=$127, glucose\,$=$527 captured accurately.\\
Arithmetic steps  & Correct & $0.016\times527=8.432$, addition correct.\\
Final answer\,(precision-aware) & \textcolor{red}{Incorrect} & $135.432\neq137.248$ under strict tolerance.\\
\midrule
\textbf{Overall}  & \textcolor{red}{\textbf{Incorrect}} & Hidden equation error \& numeric miss flagged.\\
\bottomrule
\end{tabular}
\end{center}

\subsection*{Clinical significance}
A two-point sodium underestimate may appear minor, but in critically ill, haemodynamically unstable patients, such mis-corrections can drive inappropriate fluid or insulin therapy.  
Our granular pipeline reveals both the hallucinated coefficient and the subtle numeric shortfall, preventing a misleading “pass’’ and supporting \emph{clinically defensible} deployment of LLMs.

\newpage

\section{Prompt Templates}
\label{sec:appendix-prompts}

In this appendix, we present the exact prompt templates used in our evaluation pipeline. All prompts follow a structured format consisting of a system message and a user message. For prompts related to reasoning variants such as \textbf{Direct}, \textbf{CoT}, \textbf{Oneshot}, and \textbf{Self-Refine}, please refer to our released code.

\subsection{LLM Evaluation Pipeline Prompt}

\begin{tcolorbox}[title=Prompt for LLM Eval Pipeline,
  colframe=black, colback=white,
  breakable]
\begin{lstlisting}[language=Python, breaklines=true, basicstyle=\ttfamily\footnotesize]

    def _gen_eval_prompt(self, answer, reference, name_of_step):
        # System message is the same for all steps
        system_msg = (
            "You are a medical calculation assistant. Evaluate whether each step is correct by comparing it to the gold-standard reference."
        )

        # For calculation steps, omit the gold-standard reference entirely
        if name_of_step == "calculation":
            user_msg = (
                f"{name_of_step.capitalize()} to be evaluated:\n{answer}\n\n"
                "Note: Judge ONLY the mathematical correctness of each arithmetic "
                "step (addition, subtraction, multiplication, division, powers, "
                "roots, etc.). Do NOT assess whether the formula used is appropriate "
                "or whether the input values were correct or reasonable. Treat small rounding or "
                "decimal-precision differences as acceptable"
                'Respond in this JSON format:\n\n'
                '{"result": "Correct" or "Incorrect", "explanation": "Brief justification."}'
            )
            return system_msg, user_msg

        # For all other steps, include the gold-standard reference first
        user_msg = (
            f"{name_of_step.capitalize()} to be evaluated:\n{answer}\n\n"
            f"Gold-standard reference (fully correct):\n{reference}\n\n"
            "Determine if the given part is correct according to the Gold-standard reference. "
            'Respond in this JSON format:\n\n'
            '{"result": "Correct" or "Incorrect", "explanation": "Brief justification."}'
        )

        if name_of_step == "formula":
            user_msg += (
                "\n\n"
                "Note: Judge ONLY whether the mathematical formula or scoring standard invoked is appropriate. Do NOT evaluate:"
                "the specific values plugged into the formula,"
                "the correctness of any later calculations."
                "If the gold-standard reference lists multiple valid variants (e.g., male vs. female, different ethnicities), the answer is considered correct as long as it correctly applies ANY one of those variants. If the provided formula includes more detail than the gold-standard reference but the overlapping portion is consistent and correct, it should still be considered correct."
            )
        elif name_of_step == "extracted_values":
            user_msg += (
                "\n\n"
                "Note: Check if all variables given in the gold-standard reference are found or implied. "
                "Ignore any naming discrepancies, as long as the meaning is the same. "
                "It is ok if the answer has more variables than the gold-standard reference."
                "If the given answer has a different unit than the gold-standard answer, please do conversion first. "
                "Answers with reasonable rounding errors MUST be considered Correct."
            )
        elif name_of_step in ("answer", "final_answer"):
            user_msg += (
                "\n\n"
                "Note: You ONLY need to check whether the final numerical answer matches the provided gold-standard reference. "
                "The correctness of the intermediate steps does NOT matter. If one has a unit and the other does not, please ignore the unit. "
                "If the given answer has a different unit than the gold-standard answer, please do conversion first. "
                "Answers with rounding to the nearest integer and reasonable computational deviations MUST be considered Correct."
            )

        return system_msg, user_msg

\end{lstlisting}
\end{tcolorbox}

\subsection{LLM Judge for Error Types}

\subsubsection{Formula Error}

\begin{tcolorbox}[title=Formula Error Prompts,
  colframe=black, colback=white,
  breakable]
\begin{lstlisting}[language=Python, breaklines=true, basicstyle=\ttfamily\footnotesize]

def build_formula_error_prompts(
    ground_truth_formulas: List[str],
    answers: List[str],
) -> List[Tuple[str, str]]:
    prompts = []
    for gt, ans in zip(ground_truth_formulas, answers):
        system_message = SYS_MSG.format(error_type="Formula Error")
        user_message = (
            f"Ground-Truth Formula:\n{gt}\n\n"
            f"Answer to be evaluated:\n{ans}\n\n"
            "Task: Evaluate whether the formula or scoring system used in the answer is appropriate and correctly constructed for the given clinical context.\n\n"
            "You must check for the following issues:\n"
            "- **Incorrect Formula Selection**: A completely wrong formula is used for the clinical question (e.g., using Cockcroft-Gault for AKI instead of CKD-EPI).\n"
            "- **Internal Formula Construction Errors**: The selected formula appears intended to be correct but is flawed in structure or logic. Look for:\n"
            "   Incorrect or missing coefficients or constants\n"
            "   Wrong mathematical operators (e.g., `*` instead of `^`)\n"
            "   Misused parentheses, terms in wrong places, or reversed logic\n"
            "   Hallucinated or fabricated terms in formula\n"
            "   Fabricated or omitted scoring items (e.g., omitting "recent surgery" in the Wells Score, or adding a non-existent item like "family history")\n\n"
            "**Important Notes:**\n"
            "- Do NOT evaluate variable extraction correctness here.\n"
            "- Do NOT evaluate numerical calculation or rounding accuracy.\n"
            "- If multiple formula variants exist and the answer uses any valid one, it is acceptable.\n"
            "- If the answer includes extra details but the core formula is correct, that is acceptable.\n\n"
            "Return a STRICT JSON response: "
            '{"error_present": "Yes" or "No", "explanation": ""}.'
        )
        prompts.append((system_message, user_message))
    return prompts


\end{lstlisting}
\end{tcolorbox}

\subsubsection{Variable Error}

\begin{tcolorbox}[title=Variable Error Prompts,
  colframe=black, colback=white,
  breakable]
\begin{lstlisting}[language=Python, breaklines=true, basicstyle=\ttfamily\footnotesize]


def build_variable_extraction_error_prompts(
    patient_notes: List[str],
    questions: List[str],
    ground_truth_Extracted_values: List[str],
    answers: List[str],
) -> List[Tuple[str, str]]:
    prompts = []
    for note, q, gt, ans in zip(patient_notes, questions, ground_truth_Extracted_values, answers):
        system_message = SYS_MSG.format(error_type="Incorrect Variable Extraction Error")
        user_message = (
            f"Patient Note:\n{note}\n\n"
            f"Question:\n{q}\n\n"
            f"Ground-Truth Variable Extraction:\n{gt}\n\n"
            f"Answer to be evaluated:\n{ans}\n\n"
            "Task: Determine whether the answer incorrectly extracted key variables from the patient note.\n\n"
            "You should look for the following possible errors:\n"
            "1. **Wrong value**: The extracted value (e.g., heart rate, creatinine) does not match the patient note.\n"
            "2. **Wrong unit**: The extracted unit is misinterpreted (e.g., _mol/L mistaken for mg/dL).\n"
            "3. **Wrong instance**: Multiple similar values exist (e.g., lab values from different days), and the wrong one was selected.\n\n"
            "**Do NOT evaluate:**\n"
            "- Whether the formula chosen is appropriate and correct\n"
            "- Do NOT judge whether the final answer is correct, focus only on the value extraction part.\n"
            "- Whether the numerical calculation is accurate\n\n"
            "Return a STRICT JSON response: "
            '{"error_present": "Yes" or "No", "explanation": ""}.'
        )
        prompts.append((system_message, user_message))
    return prompts


\end{lstlisting}
\end{tcolorbox}

\subsubsection{Misinterpretation Error}

\begin{tcolorbox}[title=Misinterpretation Error Prompts,
  colframe=black, colback=white,
  breakable]
\begin{lstlisting}[language=Python, breaklines=true, basicstyle=\ttfamily\footnotesize]

def build_clinical_misinterpretation_prompts(
    patient_notes: List[str],
    questions: List[str],
    ground_truth_explanations: List[str],
    answers: List[str],
) -> List[Tuple[str, str]]:
    prompts = []
    for note, q, gt, ans in zip(patient_notes, questions, ground_truth_explanations, answers):
        system_message = SYS_MSG.format(error_type="Clinical Misinterpretation Error")
        user_message = (
            f"Patient Note:\n{note}\n\n"
            f"Question:\n{q}\n\n"
            f"Scoring Rubric and corresponding result:\n{gt}\n\n"
            f"Answer to be evaluated:\n{ans}\n\n"
            "Task: Evaluate whether the clinical meaning of each finding was interpreted correctly based on the scoring rubric and patient note.\n\n"
            "This error type reflects a misunderstanding of medical knowledge or common clinical reasoning, leading to incorrect interpretation of the patient's symptoms or findings.\n\n"
            "You should check for the following types of errors:\n"
            "1. **Incorrect severity classification** (e.g., mild vs. severe ascites)\n"
            "2. **Wrong presence/absence judgment** (e.g., assigning points for recent surgery when not present)\n"
            "3. **Incorrect threshold interpretation** (e.g., age >75 incorrectly treated as <75)\n"
            "4. **Misunderstanding clinical terms or context** (e.g., interpreting 'occasional alcohol use' as 'chronic alcohol abuse')\n\n"
            "**Important Notes:**\n"
            "- The variable values may be correctly extracted from the note, but the error lies in the clinical judgment or misclassification.\n"
            "- Do NOT evaluate the correctness of the scoring formula, numeric computation, or unit conversion.\n"
            "- If the clinical inference depends on subtle wording or ambiguity in the note, highlight that in your explanation.\n\n"
            "Return a STRICT JSON response: "
            '{"error_present": "Yes" or "No", "explanation": ""}.'
        )
        prompts.append((system_message, user_message))
    return prompts


\end{lstlisting}
\end{tcolorbox}

\subsubsection{Missing Variable Error}

\begin{tcolorbox}[title=Missing Variable Error Prompts,
  colframe=black, colback=white,
  breakable]
\begin{lstlisting}[language=Python, breaklines=true, basicstyle=\ttfamily\footnotesize]

def build_missing_variable_prompts(
    patient_notes: List[str],
    questions: List[str],
    ground_truth_Extracted_values: List[str],
    answers: List[str],
) -> List[Tuple[str, str]]:
    prompts = []
    for note, q, gt, ans in zip(patient_notes, questions, ground_truth_Extracted_values, answers):
        system_message = SYS_MSG.format(error_type="Missing Variable Extraction Error")
        user_message = (
            f"Patient Note:\n{note}\n\n"
            f"Question:\n{q}\n\n"
            f"Ground-Truth Variable Extraction:\n{gt}\n\n"
            f"Answer to be evaluated:\n{ans}\n\n"
            "Task: Identify whether the answer failed to extract or include one or more variables that are necessary to perform the correct calculation.\n\n"
            "You should look for cases where:\n"
            "1. A required input variable is completely missing.\n"
            "2. The model skipped over variables because they were ambiguous or not explicitly stated.\n"
            "3. The answer proceeds with partial information, leaving out fields that the formula or score requires.\n\n"
            "**Do NOT evaluate:**\n"
            "- Whether the formula used is correct\n"
            "- Whether the extracted variables are accurate\n"
            "- Whether the final calculation is numerically correct\n\n"
            "Focus only on whether the model omitted key inputs needed to properly execute the formula or scoring rule.\n"
            "Return a STRICT JSON response: "
            '{"error_present": "Yes" or "No", "explanation": ""}.'
        )
        prompts.append((system_message, user_message))
    return prompts


\end{lstlisting}

\end{tcolorbox}

\subsubsection{Unit Error}

\begin{tcolorbox}[title=Unit Error Prompts,
  colframe=black, colback=white,
  breakable]
\begin{lstlisting}[language=Python, breaklines=true, basicstyle=\ttfamily\footnotesize]


def build_unit_conversion_error_prompts(
    patient_notes: List[str],
    questions: List[str],
    ground_truth_explanations: List[str],
    answers: List[str],
) -> List[Tuple[str, str]]:
    prompts = []
    for note, q, gt, ans in zip(patient_notes, questions, ground_truth_explanations, answers):
        system_message = SYS_MSG.format(error_type="Unit Conversion Error")
        user_message = (
            f"Patient Note:\n{note}\n\n"
            f"Question:\n{q}\n\n"
            f"Ground-Truth Explanation:\n{gt}\n\n"
            f"Answer to be evaluated:\n{ans}\n\n"
            "Task: Evaluate whether any input variable was used with the wrong unit, or skipped unit conversion when required by the formula.\n\n"
            "You should look for the following types of errors:\n"
            "1. The value is used directly without converting to the expected unit (e.g., using creatinine 134 _mol/L directly in a formula that expects mg/dL).\n"
            "2. The conversion is attempted but the result is wrong (e.g., wrong conversion factor or direction).\n"
            "3. The unit label is misunderstood or misinterpreted (e.g., confusing mEq/L with mmol/L).\n\n"
            "**Do NOT evaluate:**\n"
            "- Whether the formula chosen is appropriate\n"
            "- Whether the correct value was extracted from the note\n"
            "- Whether the afterwards numerical computation was otherwise accurate\n\n"
            "Only evaluate whether the units used match those required by the formula, and whether any necessary conversions were done correctly.\n"
            "Return a STRICT JSON response: "
            '{"error_present": "Yes" or "No", "explanation": ""}.'
        )
        prompts.append((system_message, user_message))
    return prompts


\end{lstlisting}
\end{tcolorbox}

\subsubsection{Demographic Error}

\begin{tcolorbox}[title=Demographic Error Prompts,
  colframe=black, colback=white,
  breakable]
\begin{lstlisting}[language=Python, breaklines=true, basicstyle=\ttfamily\footnotesize]


def build_adjustment_coefficient_error_prompts(
    patient_notes: List[str],
    questions: List[str],
    ground_truth_explanations: List[str],
    answers: List[str],
) -> List[Tuple[str, str]]:
    prompts = [] 
    for note, q, gt, ans in zip(patient_notes, questions, ground_truth_explanations, answers):
        system_message = SYS_MSG.format(error_type="Missing or Misused Demographic/Adjustment Coefficient Error')
        user_message = (
            f'Patient Note:\n{note}\n\n'
            f'Question:\n{q}\n\n'
            f'Ground-Truth Explanation:\n{gt}\n\n'
            f'Answer to be evaluated:\n{ans}\n\n'
            'Task: Evaluate whether demographic- or context-based adjustment coefficients were properly applied in the formula.\n\n'
            'Specifically, check for:\n'
            '1. Missing adjustment - A coefficient required by the formula is missing (e.g., sex multiplier is omitted)'
            '2. Incorrect coefficient used - The formula includes a coefficient, but it does not match the patient's characteristics (e.g., using male factor for a female patient)'
            '3. Incorrect demographic inference - The model assumes the wrong demographic category (e.g., classifying patient as non-Black when clearly stated otherwise)'
            'Common adjustment dimensions may include:'
            '- Sex (e.g., male vs. female)\n'
            '- Race/ethnicity (e.g., Black vs. non-Black)'
            '- Age thresholds\n'
            '- Pregnancy status\n'
            '- Weight class (e.g., obese vs. normal weight)'
            'Do NOT evaluate:\n'
            '- Formula structure or selection\n'
            '- Variable extraction accuracy\n'
            '- Unit conversion correctness\n'
            '- Final numerical calculation\n\n'
            'Return a STRICT JSON response: '
            '{'error_present': 'Yes' or 'No', 'explanation': ''}.'
        )
        prompts.append((system_message, user_message))
    return prompts


\end{lstlisting}

\end{tcolorbox}

\subsubsection{Arithmetic Error}

\begin{tcolorbox}[title=Arithmetic Error Prompts,
  colframe=black, colback=white,
  breakable]
\begin{lstlisting}[language=Python, breaklines=true, basicstyle=\ttfamily\footnotesize]


def build_arithmetic_error_prompts(
    answers: List[str],
) -> List[Tuple[str, str]]:
    prompts = []
    for  ans in answers:
        system_message = SYS_MSG.format(error_type="Arithmetic Error")
        user_message = (
            f"Answer to be evaluated:\n{ans}\n\n"
            "Task: All variables, units, and formula structure are assumed to be correct.\n"
            "Your task is to verify whether the **arithmetic computation** itself is correct.\n\n"
            "Check for:\n"
            "1. Basic arithmetic errors (e.g., 4 + 3 = 6)\n"
            "2. Wrong order of operations (e.g., using left-to-right instead of proper parentheses)\n"
            "3. Errors in exponentiation, multiplication, or division\n"
            "4. Missing or duplicated numeric terms\n\n"
            "**Do NOT evaluate:**\n"
            "- Formula selection or structure\n"
            "- Variable extraction\n"
            "- Unit conversion\n"
            "- Rounding or precision formatting\n\n"
            "If the calculation process is entirely accurate, a reasonable margin of error is acceptable."
            "Return a STRICT JSON response: "
            '{"error_present": "Yes" or "No", "explanation": ""}.'
        )
        prompts.append((system_message, user_message))
    return prompts


\end{lstlisting}
\end{tcolorbox}

\subsubsection{Rounding Error}

\begin{tcolorbox}[title=Rounding Prompts,
  colframe=black, colback=white,
  breakable]
\begin{lstlisting}[language=Python, breaklines=true, basicstyle=\ttfamily\footnotesize]


def build_rounding_error_prompts(
    ground_truth_explanations: List[str],
    answers: List[str],
) -> List[Tuple[str, str]]:
    prompts = []
    for gt, ans in zip(ground_truth_explanations, answers):
        system_message = SYS_MSG.format(error_type="Rounding / Precision Error")
        user_message = (
            f"Ground-Truth Explanation:\n{gt}\n\n"
            f"Answer to be evaluated:\n{ans}\n\n"
            "Task: Determine whether the numeric **final result** in the answer is imprecise due to **rounding or insufficient decimal precision**, "
            "even if the formula used and the overall arithmetic are mostly correct.\n\n"
            "This error type should be marked when rounding errors or insufficient precision in intermediate or final steps "
            "cause the final answer to fall outside the tolerance range.\n\n"
            "**Rules for evaluating precision:**\n"
            "- Use the number of decimal places in the LLM's final answer to determine the expected precision (up to a maximum of 2 decimal places).\n"
            "- If the answer is `10.65` _ round to **2 decimal places**, tolerance _0.005\n"
            "- If the answer is `10.7`  _ round to **1 decimal place**, tolerance _0.05\n"
            "- If the answer is `10.6512` _ still round to **2 decimal places**, tolerance _0.005 (overprecision beyond 2 d.p. does **not** increase accuracy expectations)\n\n"
            "**DO mark as a Rounding / Precision Error if:**\n"
            "- The calculation is mostly correct but rounding was done incorrectly (e.g., too few decimals)\n"
            "- The result deviates from the ground truth only because the final answer lacks the required precision (per above rules)\n\n"
            "**DO NOT mark as Rounding / Precision Error if:**\n"
            "- The formula used is incorrect\n"
            "- The arithmetic calculation is wrong\n"
            "- The answer is completely off due to conceptual misunderstanding\n\n"
            "Return a STRICT JSON response with:\n"
            '{"error_present": "Yes" or "No", "explanation": ""}'
        )
        prompts.append((system_message, user_message))
    return prompts



\end{lstlisting}
\end{tcolorbox}


\end{document}